\documentclass{article} 
\usepackage[final]{colm2025_conference}

\usepackage{microtype}
\usepackage{hyperref}
\usepackage{url}
\usepackage{booktabs}

\usepackage{lineno}

\definecolor{darkblue}{rgb}{0, 0, 0.5}
\hypersetup{colorlinks=true, citecolor=darkblue, linkcolor=darkblue, urlcolor=darkblue}

\usepackage{hyperref}
\usepackage{amsmath}
\usepackage{amssymb}
\usepackage{graphicx}
\usepackage{array}
\usepackage{tikz}
\usetikzlibrary{patterns}
\usetikzlibrary{positioning}
\usepackage{listings}
\usepackage{xspace}
\usepackage{caption}
\usepackage{multirow}
\usepackage[capitalise]{cleveref}
\usepackage{wrapfig}
\usepackage{makecell}

\newcommand{\md}[2]{\in \mathbb{R}^{#1 \times #2}}
\newcommand{\oursfull}{Multi-Token Attention}
\newcommand{\ours}{MTA}
\newcommand{\std}[1]{\small (#1)}

\newcommand{\stddev}[1]{\scriptsize $\pm$ #1}

\title{
Multi-Token Attention 
}{}

\author{Olga Golovneva
  \quad
  Tianlu Wang 
  \quad
  Jason Weston 
  \quad
  Sainbayar Sukhbaatar \\ \\
  FAIR at Meta
}

\begin{document}

\ifcolmsubmission
\linenumbers
\fi

\maketitle

\begin{abstract}
Soft attention is a critical mechanism powering LLMs to locate relevant parts within a given context.
However, individual attention weights are determined by the similarity of only a single query and key token vector.
This ``single token attention'' bottlenecks the amount of information used in distinguishing a relevant part from the rest of the context. 
To address this issue, we propose a new attention method, 
\emph{\oursfull{}} (\ours{}), 
which allows LLMs to condition their attention weights on multiple query and key vectors simultaneously. 
This is achieved by applying convolution operations 
over queries, keys and heads, allowing nearby queries and keys to affect each other's attention weights for more precise attention.
As a result, our method can locate relevant context using richer, more nuanced information that can exceed a single vector's capacity.
Through extensive evaluations, we demonstrate that \ours{} achieves enhanced performance on a range of popular benchmarks. 
Notably, it outperforms Transformer baseline models on standard language modeling tasks, and on
tasks that require searching for information within long contexts, where our method's ability to leverage richer information proves particularly beneficial\footnote{Code is available at \href{https://github.com/facebookresearch/RAM/tree/main/projects/mta}{https://github.com/facebookresearch/RAM/tree/main/projects/mta}}.
\end{abstract}

\section{Introduction}
The attention mechanism \citep{bahdanau2014neural,Vaswani+2017}
is a critical component of 
Large Language Models (LLMs) that enables them to retrieve and combine information from different parts of the context.
%
Attention is especially useful when the context contains a large number of tokens, as focusing on the relevant part while disregarding distractions becomes crucial. However, numerous works have shown that standard attention can suffer from suboptimal performance in this setting \citep{Kamradt2023Needle,kuratov2025babilong}.

Standard multi-head attention works by comparing the similarity of the current query vector and the key vectors corresponding to the context tokens using their dot products.
The keys similar to the query obtain higher attention weights, and subsequently their value vectors dominate the output vector.
For example, a query vector corresponding to a token ``Alice'' is capable of locating all mentions of ``Alice'' in the context.
However, each attention weight  conditions only on a single key and query vector (besides the normalization to sum to one).

We argue that the dependency on single token vector similarity brings a fundamental limitation to the attention mechanism.
In many cases, the relevant part of the context cannot be identified by a single token.
For example, looking up a sentence that mentions both ``Alice'' and ``rabbit'' requires the query vector to encode both tokens.
Looking up ``Alice'' with one attention head 
and using another head for ``rabbit'' could find their mentions separately, but is not sufficient to pinpoint where both are mentioned together.
While it is possible to encode multiple tokens into a single vector via the layers of the Transformer, this requires increased dimension, and for the model to use lots of its capacity for this task.

In this paper, we propose a novel attention mechanism that goes beyond this ``single token'' bottleneck, which we call \oursfull{} (\ours{}).
The high level goal is to make it possible to use the similarities of \emph{multiple vector pairs} to determine where attention must focus.
We achieve this by making straight-forward modifications to the existing attention mechanism.
In particular, we design convolution operations over attention weights that operate on three dimensions: keys, queries, and attention heads.
This allows its attention weights to condition on neighboring keys, previous queries, and other heads.
Intuitively, following our previous example, \ours{} can find mentions of ``Alice'' and ``rabbit'' separately first, and then combine those attentions together to focus only on where both exist.

We first experiment with a motivating toy task that reveals the shortcoming of standard attention and demonstrate that \ours{} can easily solve it. 
Next, we test our method at scale by pre-training 880M parameters models on 105B tokens on a standard language modeling task.
There we see \ours{} bring improvements in terms in validation perplexity as well as standard benchmark tasks, while only increasing the number of parameters by $0.001\%$.
Further, we evaluate the resulting models on long-context tasks such as Needle-in-the-Haystack and BabiLong where \ours{} outperforms the baselines by a significant margin.
Finally, we investigate scaling laws of the proposed model and compare them with baselines.

\section{Background on multi-head attention}
\begin{figure}[t]
\vspace{-7mm}
    \centering
    \includegraphics[width=0.72\linewidth,clip]{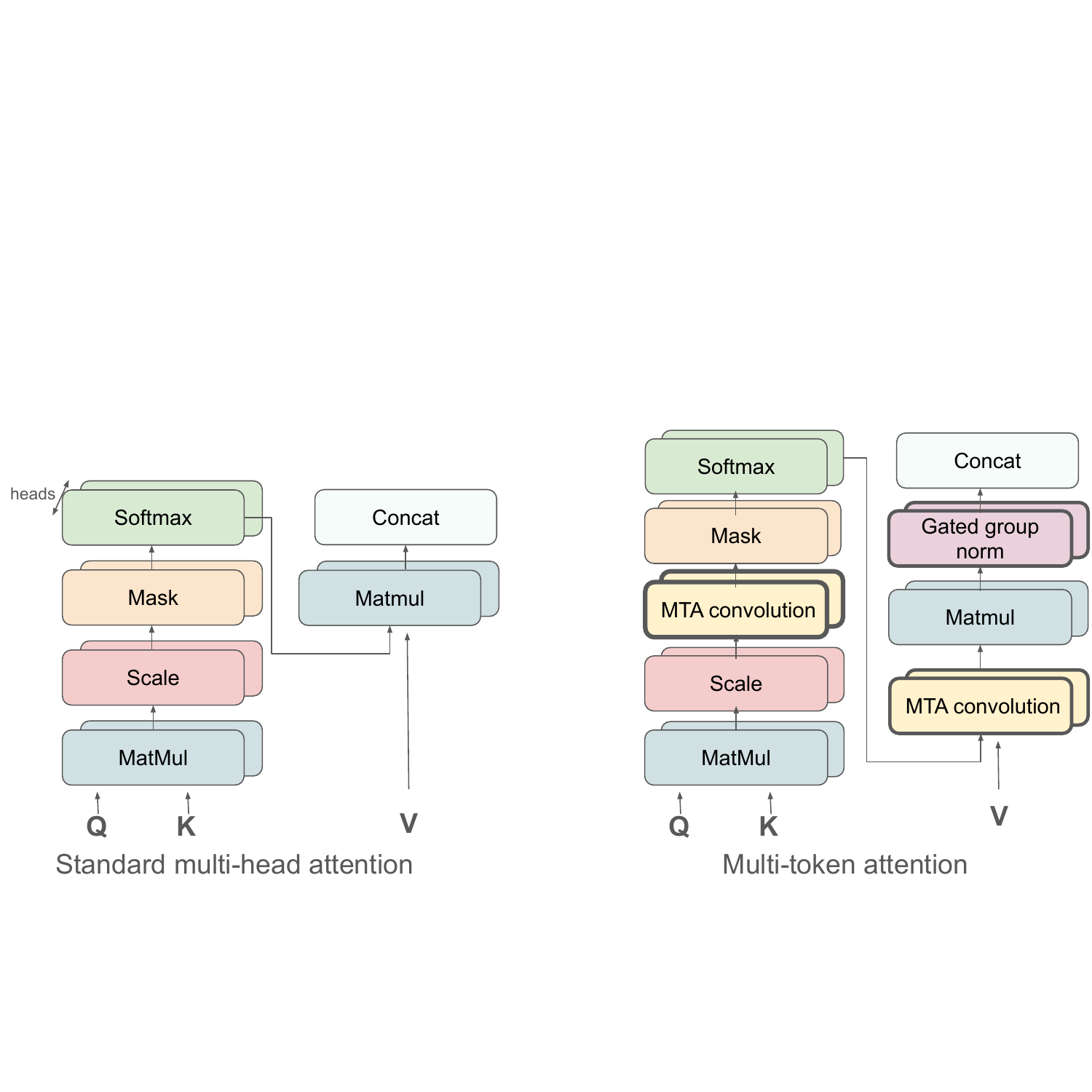}

    \caption{{\bf \oursfull~(\ours{})} (right), compared to standard attention (left). In \ours{}, within each head we apply a key-query convolution on the attention scores and a head convolution across groups of heads. We repeat this operation after the softmax. Finally, we apply a group normalization with scalar gating before final concatenation.}
    \label{fig:attn_schema}
\end{figure}

We first describe the standard 
multi-head attention  \citep{Vaswani+2017} and define the notation we use. In decoder-only Transformer architectures, the model receives a sequence of tokens $[x_1,...,x_T]$ of length $T$ as an input.
These tokens then undergo a series of transformations into hidden states $H=[h_1, \ldots h_T]^\top \in \mathbb{R}^{T \times D}$ through an embedding layer and  repeated  transformer layers.
Each layer consists of a multi-head attention submodule, a feedforward network, and normalization operations. Multi-head attention includes $M$ heads with dimensions $d=D/M$ working in parallel. Each head uses key, value and query projections $W_k, W_v, W_q \in \mathbb{R}^{D \times d}$
to construct key, value and query vectors:
    \[
        K = H W_k , \quad V = H W_v, \quad Q = H W_q 
    \]
The attention logits $\hat{A}$ and weights $A$ (i.e.\ probabilities) are then calculated as follows:
\begin{equation}
\hat{A}={QK^\top}/{\sqrt{d}}, \quad A = \text{Softmax}( \text{Mask}_{-\infty}(\hat{A}) ),
\label{eq:attn}
\end{equation}
where the softmax operates over the key dimension, and the mask function replaces values at $(i,j)$ with $-\infty$ for $i<j$ to prevent information leaking from future tokens.
Finally, the attention output $A V \md{T}{d}$ from all heads is then concatenated, and multiplied by the output projection $W_o \md{D}{D}$ which is then passed to the normalization and feedforward components.
This standard  approach is summarized in  \autoref{fig:attn_schema} (left).

\section{\oursfull}

Each attention value in standard multi-head attention, see \autoref{eq:attn}, depends solely on a single key and query vector. 
That means all the necessary information for finding and attending to a relevant part of the context must be compressed into these single vectors. 
This might not be ideal if we are looking for a sentence containing multiple elements. Consider for example the sentence ``Where did Alice see the rabbit?''.
We could try to find instances of ``Alice'' and ``rabbit'' independently and then check if there is a sentence that has both.
Let $q_a$ and $q_r$ be query vectors encoding ``Alice'' and ``rabbit'' respectively (assuming a word tokenizer), then their attention weights are computed as follows:
\begin{align}
a_a = \text{Softmax}(q_a K^\top /{\sqrt{d}}), \quad
a_r = \text{Softmax}(q_r K^\top /{\sqrt{d}})    
\end{align}
By doing normal attention with those queries, we can attend where ``Alice'' and ``rabbit'' are mentioned in the context.
All we have to do then is to check if both attention weights $a_a$ and $a_r$ have higher probabilities at the same nearby locations, e.g. in the same sentence, which will indicate that the sentence mentions both ``Alice'' and ``rabbit''.
Unfortunately, normal attention lacks such interaction between attention maps, and instead only uses them to compute output values.
Even if we use different attention heads to find ``Alice'' and ``rabbit'', there is no mechanism to combine these attention weights.
This motivates us to modify the attention mechanism to allow combining different attention maps from nearby locations (both in terms of query and key locations), or between different attention heads.

As shown in \autoref{fig:attn_schema} (right),  our proposed \oursfull~consists of three important components built on top of multi-head attention: key-query convolution, head mixing convolution, and group normalization with gating mechanism. The overall \ours{} convolution applies the key-query convolution to combine multiple keys and queries within heads, and the head convolution to share knowledge between heads and amplify important information. Finally, we apply group normalization with scalar gating to push back against residual streams and improve gradient flow.
In this section, we will describe each component of \ours{} in detail. 

\begin{figure}[t]
\vspace{-7mm}
    \centering
    \includegraphics[width=0.7\linewidth,clip]{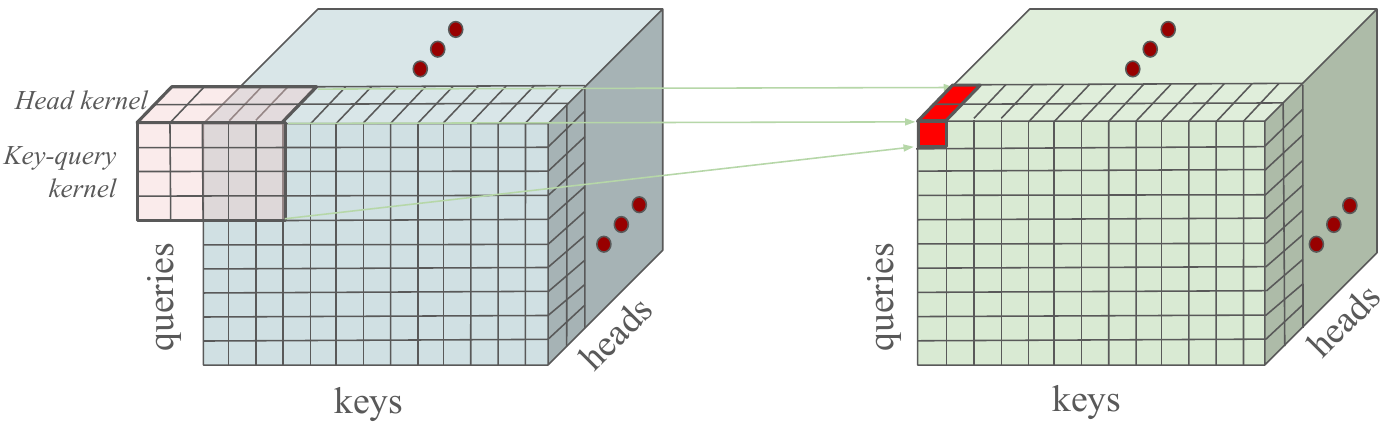}
    \caption{\ours{} applies key-query and head convolution over attention values.}
    \label{fig:kernel_schema}
\end{figure}

\subsection{Key-query convolution}
\label{sec:conv}

\paragraph{Pre-softmax convolution}
\ours{} applies a convolution operation on attention logits to combine information from multiple query and key tokens:
\begin{equation}
A = \text{Softmax}\left( \text{Conv2d}_\theta(\hat{A}) \right),
\end{equation}
where $\text{Conv2d}_\theta$ is a 2-dimensional convolution operation with kernel weights $\theta$, and kernel sizes $(c_q, c_k)$. 
Convolution is applied to the key and query length dimensions, while the batch and head dimensions remain independent.
More precisely, the attention weight $a_{ij}$ from query $q_i$ to key $k_j$ is computed as follows:
\begin{equation}
    \label{eq:attn_ij}
    a_{ij} = \text{Softmax} \left( \sum_{i'=0}^{c_q-1} \ \sum_{j'=-\lfloor{c_k/2}\rfloor}^{\lceil{c_k/2}\rceil-1} \ \mathbf{1}_{i \ge j-j'} \ \theta_{i',j'} \ q_{i-i'} \ k_{j-j'}^\top \  / \ {\sqrt{d}} \right)
\end{equation}
Given that the query position is $i$, any information from position $> i$ should not be used to prevent information leakage from  the future.
That is why only past queries are used in \autoref{eq:attn_ij}.
For keys, we use indicator function $\mathbf{1}_{i \ge j-j'}$ to zero out future keys.
However such masking is too complex to implement (it is necessary to modify the convolution cuda kernels), thus we propose a simpler version that applies the existing causal masking twice:
\begin{equation}\label{eq:kq_attn}
A = \text{Softmax}\left( \text{Mask}_{-\infty} \left( \text{Conv2d}_\theta \left( \text{Mask}_0(\hat{A}) \right) \right) \right).
\end{equation}
Here, the first masking uses 0 so that those values will not affect the output from the convolution.
Although this version masks out a little more than necessary, it is simpler to implement and prevents information leakage, so we use it as our default.

\paragraph{Post-softmax convolution}
Similarly, we can apply the convolution on top of the attention weights instead of the logits 
\begin{equation}
A =  \text{Mask}_0\left( \text{Conv2d}_\theta \left( \text{Softmax}\left( \text{Mask}_{-\infty}( \hat{A})   \right) \right) \right) .
\end{equation}
This makes interaction between attention weights additive instead of multiplicative. We will experiment with both versions, but will use the pre-softmax version by default.

Each attention head has separate $\theta$ parameters, so they can perform different convolution operations.
The chosen kernel dimensions dictate how far away tokens can be combined together.
In the above example, the question ``Where did Alice see the rabbit?'' will make $q_a$ and $q_r$ queries separated by two tokens (assuming a word tokenizer), so we need $c_q=4$ to cover both queries.
Similarly, the target sentence ``Alice saw the white rabbit under the tree'' for example produces keys $k_a$ and $k_r$ separated by three tokens, so $c_k=5$ is sufficient for combining them.
Before applying the convolution, we pad the input with an appropriate amount of zeros so that each convolution operation has a valid input.

\subsection{Head mixing convolution}
Unlike the key-query convolution, 
which allows mixing of attention weights from different time steps,
we further propose to use a head convolution over groups of heads, 
so attention weights from different heads can be combined.
In particular, for a head convolution kernel of size $c_h$, all heads are divided in $M/c_h$ groups. Within each group, we apply a non-overlapping convolution operation\footnote{Since the kernel size $c_h$ and group size $c_h$ are the same, it can also be viewed as fully-connected.}. This way, \ours{} allows not only to condition attention weights on multiple query and key vectors within each head, but also shares attention information across heads.
For example, consider splitting all heads into groups of two, such that the kernel size is $c_h=2$.
Let us use superscript to denote head indices, so  $A^1$ and $A^2$ are attention weights from two different heads. Then the new attention weights are:
\begin{equation}\label{eq:head_mix}
    A_\text{new}^1 = w_{11} A^1 + w_{12} A^2 , \quad A_\text{new}^2 = w_{21} A^1 + w_{22} A^2 ,
\end{equation}
where $w_{11},w_{12},w_{21},w_{22}$ are kernel weights.
Here the mixing occurred after the softmax, but we can also mix logits before the softmax:
\begin{align*}
\hat{A}_\text{new}^1 = w_{11} \hat{A}^1 + w_{12} \hat{A}^2 , \quad \hat{A}_\text{new}^2 = w_{21} \hat{A}^1 + w_{22} \hat{A}^2 ,
\end{align*}
Like the key-query convolution, we experiment with both versions.




\subsection{Putting everything together}
In the previous sections, we introduced two different ways of mixing attention weights, one across key-query time steps and another across different heads.
These two can be implemented together in a single \ours{} module.
Because each has pre and post-softmax versions, there are multiple different ways of combining them.

If both mixing methods are pre-softmax, then they can be implemented by a single 3-dim convolution operation as shown in \autoref{fig:kernel_schema}.
Two of these dimensions will span key and query dimensions as described in \autoref{sec:conv}.
The third dimension will be across attention heads, but on groups of $c_h$ heads. 
The same 3-dim convolution can be applied after softmax to perform both mixing methods post-softmax.
The third possibility is to apply the key-query convolution before softmax, and then head mixing after softmax.
This is also straightforward by applying \autoref{eq:head_mix} on the attention weights obtained from \autoref{eq:kq_attn}.

Finally, we follow \citet{ye2024differential} who also apply normalization, but in our case we replace the layer-dependent scaling with a sigmoid gating mechanism. Sigmoid gating allows the model to switch heads on and off across layers and better adapt to different tasks.
We also perform ablations across different normalization approaches.

\section{Experiments}
We conduct experiments with the \ours{} architecture, comparing to several baselines on a set of standard and long-range dependency tasks, starting with a toy task.
Unless otherwise specified, we apply the \ours{} convolution both  pre-softmax and post-softmax.

\subsection{Motivating toy task}
\begin{table}[t]
\vspace{-7mm}
    \centering \small
    \begin{tabular}{lcccccc}
        \toprule
        & \multicolumn{3}{c}{\it Block size $N=5$} & \multicolumn{3}{c}{\it Block size $N=8$}\\
        \cmidrule(lr){2-4}\cmidrule(lr){5-7}
        \bf Model & \bf All & \bf First & \bf Last  & \bf All & \bf First & \bf Last \\
        \midrule
        Transformer & 51.6 \stddev{43.1} & 78.2 \stddev{1.5} & 1.3 \stddev{0.4} & 31.2 \stddev{50.3} & 28.9 \stddev{24.9} & 58.4 \stddev{46.8} \\ 
        MTA & 0.1 \stddev{0.1} & \phantom{0}0.0 \stddev{0.0} & 0.0 \stddev{0.0} & 0.1 \stddev{0.0} & 0.0 \stddev{0.0} & 0.0 \stddev{0.0}\\
        \bottomrule
    \end{tabular}
    \caption{\textbf{Error rates (\%) on the motivating toy task} where the model needs to locate a block of letters containing the given $L=2$ letters. The output should contain \emph{all}, \emph{first}, or \emph{last} tokens of the target block. \ours{} perfectly solves this task while a standard Transformer struggles to learn it. We report average error rate over 3 seeds and their standard deviations.}
    \label{tab:toy}
\end{table}

We start with a simple toy task to demonstrate the effectiveness of our method over standard multi-head attention.
In this task, the model is given a sequence of blocks where each block consists of  $N$ random letters.
This is followed by $L$ question letters ($L<N$).
The objective is to find the block that contains all question letters in any order.
Then the model must output
\emph{all} letters of the target block, or only its \emph{first} or \emph{last} token (as three separate task variants).

Despite its simplicity, this task poses a challenge to standard attention because it requires $L$ pieces of information (i.e. question letters) to identify the target block.
To succeed, standard soft attention must encode all $L$ question letters into a single query vector.
In contrast, \ours{} can first find the locations of each question letter, then use its convolution operation to increase the attention to the locations where all $L$ letters are found together.

For this task, we train a small Decoder-only Transformer model with 4 layers, 2 heads and 256 hidden dimensions. The results are shown in \autoref{tab:toy}. As expected, Transformer with standard multi-head attention struggles to solve this task, often completely failing to find the target blocks even though the questions have only $L=2$ letters.
This highlights the inherent limitation of standard attention conditioned on single token vectors.
For \ours{}, we set the query kernel size $c_q=2$ to match the number of question letters, and the key kernel $c_k=2N-1$ so it can cover a whole block on both sides.
To simplify the experiments, the key-query convolution is only applied before softmax, and no head convolution is used.
This way, it is sufficient for each query vector to encode only a single question letter while the convolution operation can aggregate their locations to find the target block.
As a result, \ours{} successfully solves all the versions of the task with near zero error rate. See \autoref{sec:toy_detail} for more training details.

\subsection{Large language modeling}
For language modeling experiments, we perform pre-training of 880M-size models and compare the Transformer model \citep{Vaswani+2017}, Differential Transformer (\textsc{Diff} Transformer) \citep{ye2024differential}, Transformer with Talking heads attention \citep{shazeer2020talking}, and Transformer with \ours. \textsc{Diff} Transformer calculates attention scores as the difference between two separate softmax attention maps, while Talking heads attention applies linear projections across heads before and after softmax operations, 
thus being related to our head convolution approach, so we use them as baselines.
All models are trained in the same setup on the SlimPajama \citep{cerebras2023slimpajama} dataset for 105B tokens using the Lingua framework \citep{meta_lingua}. 
Training details are provided in \autoref{sec:trainig_details}. To improve training efficiency, we apply the key-query convolution on every 4th layer, while head convolution is applied on all layers. Kernel dimensions are fixed at $c_q=6, c_k=11$, and we mix groups of $c_h=16$ heads within each layer. We ablate these parameters in \autoref{sec:ablations}.

For each model, we conduct training twice and report average validation perplexity in \autoref{tab:valid-ppl}. We observe consistent improvements across all validation datasets for the model trained with \ours{}, even though the key-query convolution was only applied to a quarter of the layers, 
keeping the total number of parameters on-par with the Transformer baseline (see Appendix \autoref{tab:extra-param}). We also note that group normalization with layer scaling is an important component that drives superior performance for both the \textsc{Diff} Transformer and \ours{} architectures. 

\begin{table*}[t]
\vspace{-7mm}
  \centering
  \small
\begin{tabular}{l|cccccccc}
\toprule
\bf Model & \bf arxiv & \bf book & \bf c4 & \bf cc & \bf github & \bf se & \bf wiki & \bf Avg PPL $\downarrow$ \\
\midrule
\hspace{-2mm}\textit{Pretraining} \\
Transformer & 4.65 & 13.47 & 20.20 & 14.41 & 4.28 & 10.13 & 11.64 & 11.25 \std{0.00} \\
\textsc{Diff} transformer & 4.62 & 13.33 & 19.99 & 14.28 & 4.25 & 10.04 & 11.54 & 11.15 \std{0.02} \\
Talking heads & 4.59 & 13.26 & 19.82 & 14.15 & 4.20 & 9.93 & 11.33 & 11.04 \std{0.00} \\
MTA & \bf 4.54 & \bf 13.09 & \bf 19.63 & \bf 14.00 & \bf 4.12 & \bf 9.76 & \bf 11.18 & \bf 10.91 \std{0.01} \\
\midrule
\hspace{-2mm}\textit{Long context finetuning} \\
Transformer & 4.32 & 13.18 & 20.14 & 14.08 & 3.96 & 9.84 & 11.63 & 11.02 \\
\textsc{Diff} transformer & 4.28 & 13.01 & 19.87 & 13.93 & 3.90 & 9.72 & 11.49 & 10.89 \\
Talking heads & 4.29 & 13.25 & 19.95 & 13.90 & 3.86 & 9.66 & 11.29 & 10.88 \\
MTA & \bf 4.21 & \bf 12.77 & \bf 19.51 & \bf 13.66 & \bf 3.79 & \bf 9.46 & \bf 11.14 & \bf 10.65 \\
\bottomrule
\end{tabular}
\caption{Validation perplexity for 880M Transformer model on SlimPajama dataset after training for 105B tokens, and finetuning with $2048\rightarrow 4096$ context extension for another 10.5B tokens. Pretraining perplexity was averaged across two runs.}
\label{tab:valid-ppl}
\end{table*}

We further evaluate our models on a set of popular benchmarks in a zero-shot setup as shown in \autoref{tab:baseline_tasks}. The model trained with \ours{} outperforms the baselines on most of them and achieves a higher average score, despite these not being long-context tasks.

\begin{table*}[t]
  \centering
  \resizebox{\textwidth}{!}{
  \small
  \setlength{\tabcolsep}{3pt}
\begin{tabular}{lrcccccccccc}
        \toprule
 \bf Model &
           {\small{\textbf{BoolQ}}} &	
           {\small{\textbf{PIQA}}} & 
           {\small{\textbf{SIQA}}} & 
           {\small{\textbf{HellaS}}} &
           {\small{\textbf{WinoG}}} &
           {\small{\textbf{ARCe}}} &
           {\small{\textbf{ARCc}}} &
           {\small{\textbf{OBQA}}} &
           {\small{\textbf{ MMLU}}} & 
           {\small{\textbf{~Avg $\uparrow$}}}\\

\midrule
Transformer & 56.2& 70.2 & 39.9 & 38.5 & 56.4 & 57.9 & \bf 25.9 & \bf 23.8 & 24.5 & 43.7 \std{0.3} \\
\textsc{Diff}  transformer & 59.6 & 70.5 & 39.7 & 38.9 & 56.4 & 57.7 & 25.6 & 21.4 & 24.9 & 43.9 \std{0.5} \\\
Talking heads & 61.9 & 71.4 & \bf 40.6 & 39.4 & 54.5 & 58.2 & 25.2 & 23.6 & 24.7 & 44.4 \\
MTA & \bf 62.1 & \bf 71.7 & 40.4 & \bf 39.7 & \bf 57.2 & \bf 58.9 & 24.7 & 23.6 & \bf 25.8 & \bf 44.9 \\
\bottomrule
\end{tabular}
}
\caption{Pretrained models' evaluation results on standard benchmarks. Results are averaged over two model training runs for each method.}
\label{tab:baseline_tasks}
\end{table*}

\subsection{Long context finetuning}
\label{sec:long-cont}

We further finetune our models on the same mix of datasets for another 10.5B tokens, but increase the context length from 2048 to 4096. We increase RoPE's theta to $500000$, change the weight decay to $0$, and reduce warm up steps to $50$. Other parameters remain the same as during pretraining (see \autoref{sec:trainig_details}). The resulting Transformer model with \ours{} similarly outperforms the new baselines in perplexity evaluations as shown in \autoref{tab:valid-ppl}.

\subsection{Long-range dependency tasks}

It was previously shown that Transformers struggle to find relevant information especially in the middle of long context \citep{liu2024lost, liu2025thus}.
To test \ours{} in this setting, we evaluate trained models on three tasks: \textsc{Lambada} \citep{paperno2016lambada,radford2019language}, Needle-In-A-Haystack \citep{Kamradt2023Needle} and BabiLong \citep{kuratov2025babilong}. All these tasks require models to almost sharply attend to the long-range tokens buried in the context.

\textbf{\textsc{Lambada} } is a collection of texts that test the model's ability to search long-range dependencies in the context. In particular, this dataset is designed such that humans can correctly predict the next word only if they have the whole text, but not if they only see the last sentence preceding the target word. We observe models trained with \ours{} are better at correctly guessing the next word (\autoref{tab:lambada}), significantly outperforming the baseline Transformer model.

\begin{table*}[t!]
\vspace{-7mm}
  \centering \small
\begin{tabular}{lcc}
        \toprule
 \bf Model & \bf \textsc{Lambada} standard  $\downarrow$  & \bf \textsc{Lambada} OpenAI  $\downarrow$ \\

\midrule
Transformer & 17.6 & 9.5 \\
\textsc{Diff} transformer & 14.9 & 9.3 \\
Talking heads & 15.1 & 8.9 \\
MTA & \bf 13.2 & \bf 8.4 \\

\bottomrule
\end{tabular}
\caption{ Perplexity evaluations on the \textsc{Lambada} standard \citep{paperno2016lambada} and \textsc{Lambada} OpenAI \citep{radford2019language} datasets.}
\label{tab:lambada}
\end{table*}

\begin{figure}[t]
    \centering
    \includegraphics[width=0.5\linewidth]{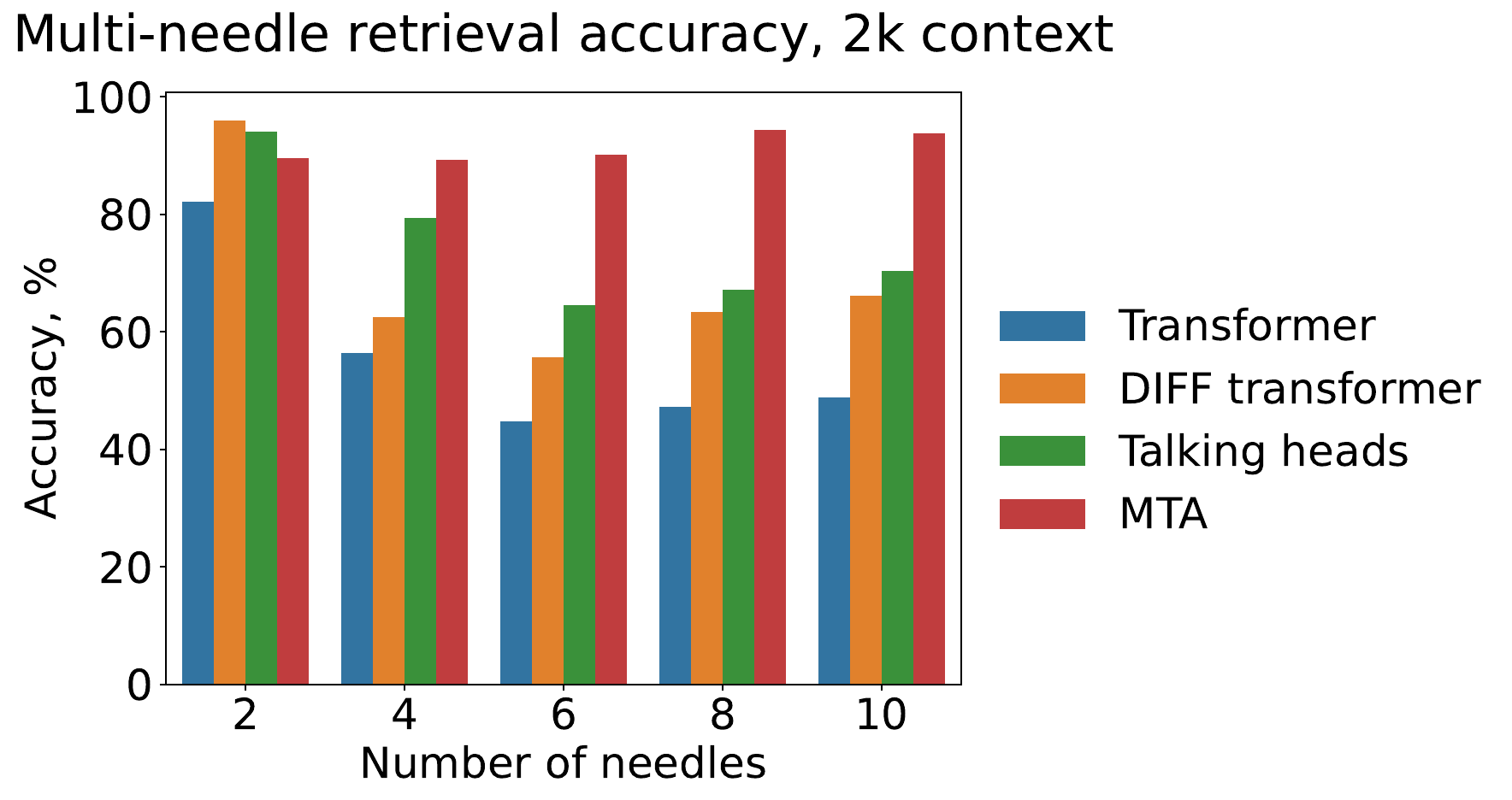}\includegraphics[width=0.5\linewidth]{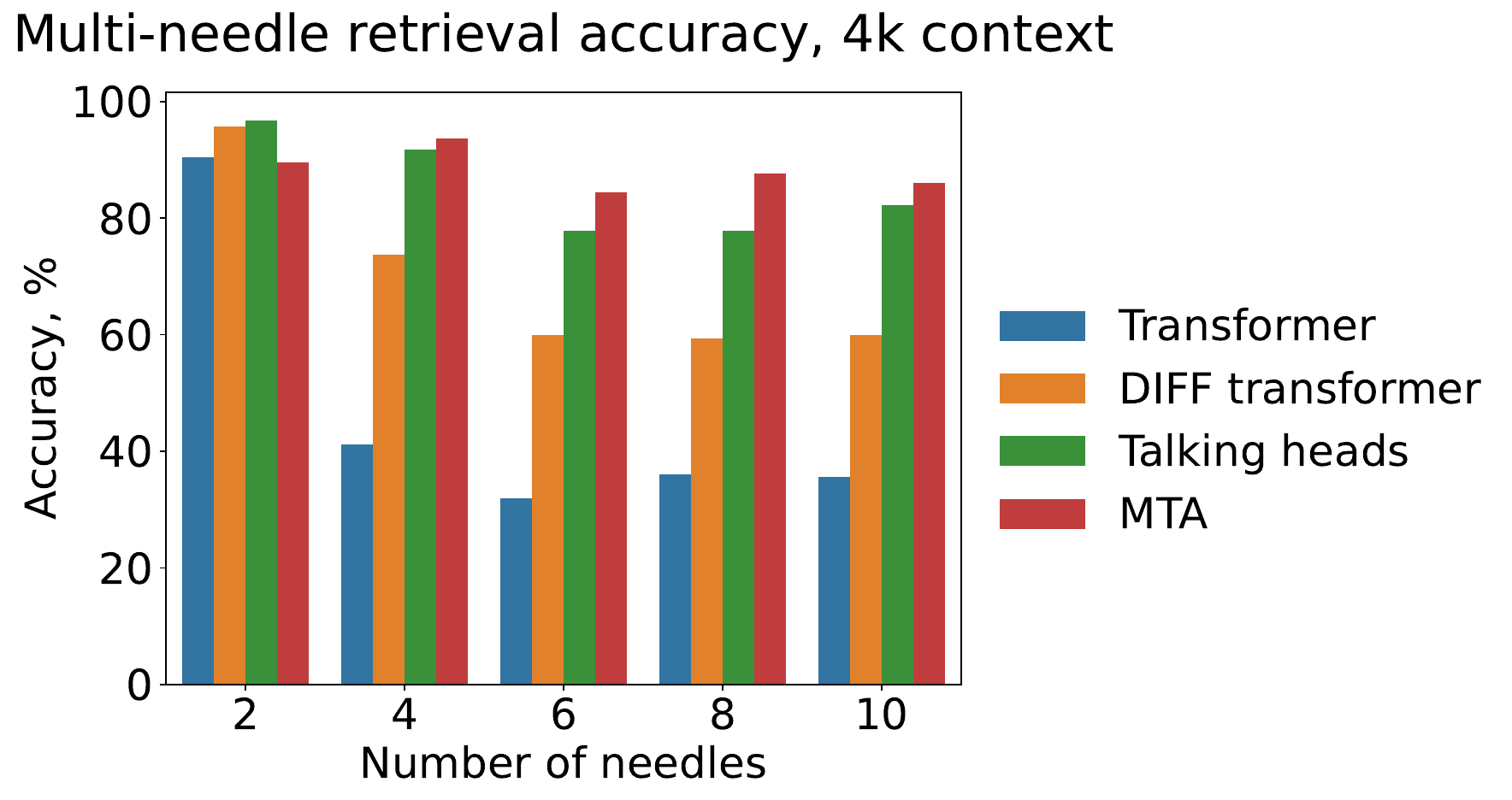}
    \caption{Multi-needle retrieval accuracy 
averaged over different needle insertion depths. Pretrained models are evaluated with 2K
context (left), while finetuned models are evaluated with 4K context (right).
}
    \label{fig:needles_avg}
\end{figure}

\textbf{Needle-In-A-Haystack} We probe our models by inserting 2, 4, and 6 needles in 2k and 4k context windows at varying depths. We additionally ensure that for each new sample, the needles are shuffled, thus removing bias the models might have toward extracting the needle that was inserted first or last. Each setup is evaluated on 500 samples. Accuracy evaluations are averaged across the depths of insertion. As reported in \autoref{fig:needles_avg}, we observe a significant improvement in needle extraction abilities for models trained with \ours{} across all needle counts and varying context lengths. Breakdown by depth is reported in \autoref{app:needle}.

\begin{figure}[t]
\vspace{-7mm}
    \centering
    \includegraphics[width=0.41\linewidth]{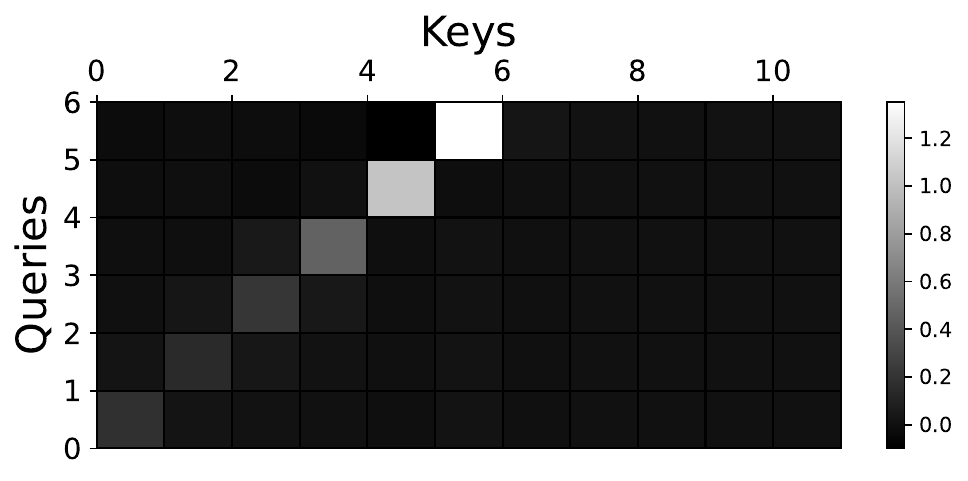}
    \includegraphics[width=0.48\linewidth]{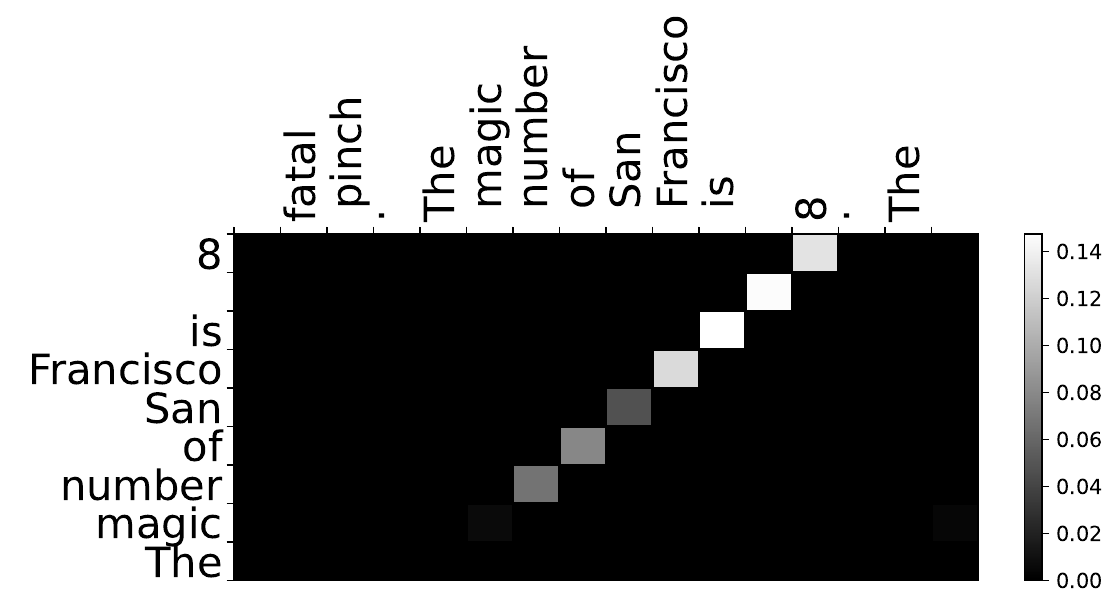}
    \caption{Kernel pattern (left) and corresponding attention map (right), which has the highest attention scores on the targeted needle "The magic number of San Francisco is 8".
    This kernel amplifies attention if a query token sequence matches a key sequence -- 
useful for searching for related sentences that match the current sentence.
}
    \label{fig:kernels_maps}
\end{figure}

\textbf{BabiLong} This benchmark tries to to assess a language models' ability to reason across facts scattered in long documents. We focus on tasks QA1-5 \citep{weston2015towards} where a correct response requires various numbers of facts or argument relations. Examples of input and target output can be found in~\autoref{tab:babilong_examples}.  We present average accuracy in~\autoref{fig:qa4_5}(left) and  per-task in Appendix~\autoref{fig:qa1_3}. We observe that the \ours{} model performs well compared to other models, especially when there is more distraction text (4K tokens) in the input. 

\begin{figure}[t]
    \centering
    \hfill
    \includegraphics[height=4.9cm]{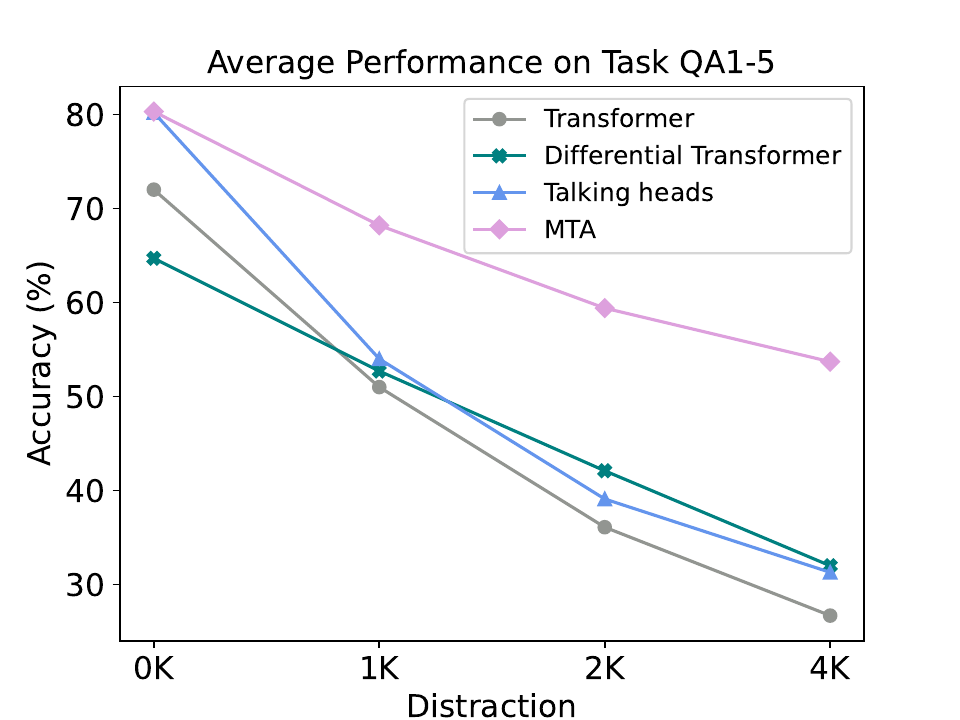}
    \hfill
    \includegraphics[height=4.43cm]{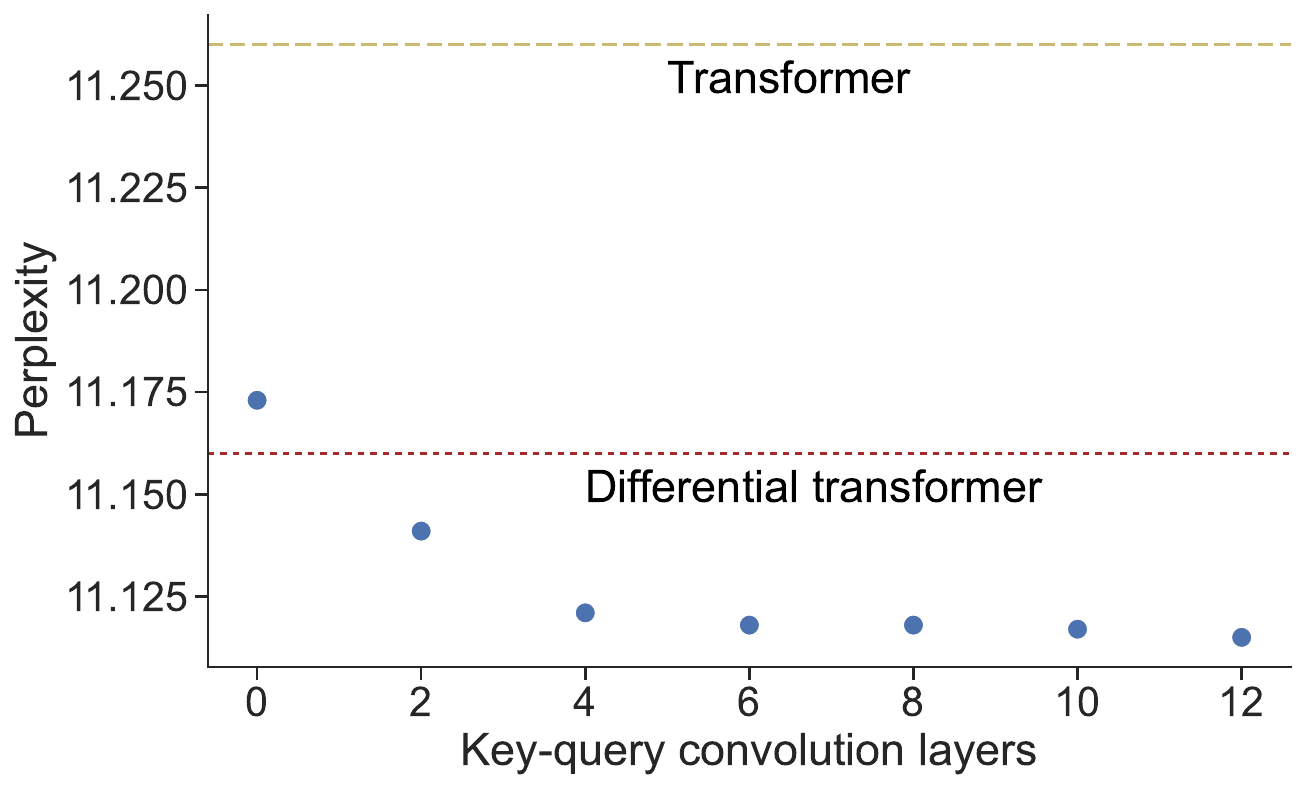}
    \hfill
    \caption{\textbf{(left)} Average accuracy on QA1-5 tasks in BabiLong. The models are all pretrained with 2K context and then finetuned with 4K context. Distraction text length varies from 0K (no distraction) to 4K tokens. \ours{} consistently outperforms the baseline models. \textbf{(right)} Ablation on the number of \ours{} layers with key-query convolutions (head convolution is applied to all layers). We report average validation perplexity on SlimPajama.}
    \label{fig:qa4_5}
\end{figure}




\subsection{Kernel patterns}

Key-query and head convolution kernels across different layers and heads are shown in \autoref{app:kernel_patterns}. We observe that a number of 
of the key-query kernels are close to identity, with near-one weight learned for targeted query and key, and near-zero weights learned for all other positions. 
However, there are numerous kernels that are more intricate.

One such kernel is shown in \autoref{fig:kernels_maps}(left), which has a diagonal structure.
This kernel will amplify attention if a sequence of query tokens match a sequence of keys.
Such ability is useful in searching sentences that matches the current sentence.
Indeed, we observe that this particular kernel focuses attention on the target needle (see \autoref{fig:kernels_maps}(right)) in the Needle-in-the-Haystack task, where we must locate the magic number of the matching city.


Other kernels, for example, can be viewed as priming, amplifying if the same key was attended by previous queries (head 5 on \autoref{fig:kern_l7}), 
or edge detecting, amplifying the first or last of multiple contiguous keys with high attention (heads 8 and 11 on \autoref{fig:kern_l15}). We leave further exploration of key-query kernel maps and their meaning for future research. 

Head kernel patterns, on the contrary, are simpler due to their small size, as shown in Appendix \autoref{fig:head_kern}. 
Besides an identity with scaling, a common pattern is contrasting: subtracting one attention weight from another.
We also observe that the kernel scales increase with layers when there is no group normalization (see Appendix \autoref{fig:head_diag}).
This is probably to compete with the residual stream, which gets larger with the model's depth.
However, this pattern is not present with group normalization because it will undo the effect of such scaling.

\subsection{Ablations}
\label{sec:ablations}


\paragraph{Key-query convolution} To understand how key-query convolution affects the model performance, we run ablation studies on the number of layers where key-query convolution is added to the attention. The results shown in \autoref{fig:qa4_5}(right) demonstrate that when only 2 layers are enhanced with \ours{}, the model can outperform strong baselines, while 6 layers with \ours{} strikes a balance between performance and additional complexity (see \autoref{sec:compute} for more about computational complexity).

\begin{figure}[t]
    \centering
    \vspace{-7mm}
    \hfill
    \includegraphics[height=4.5cm]{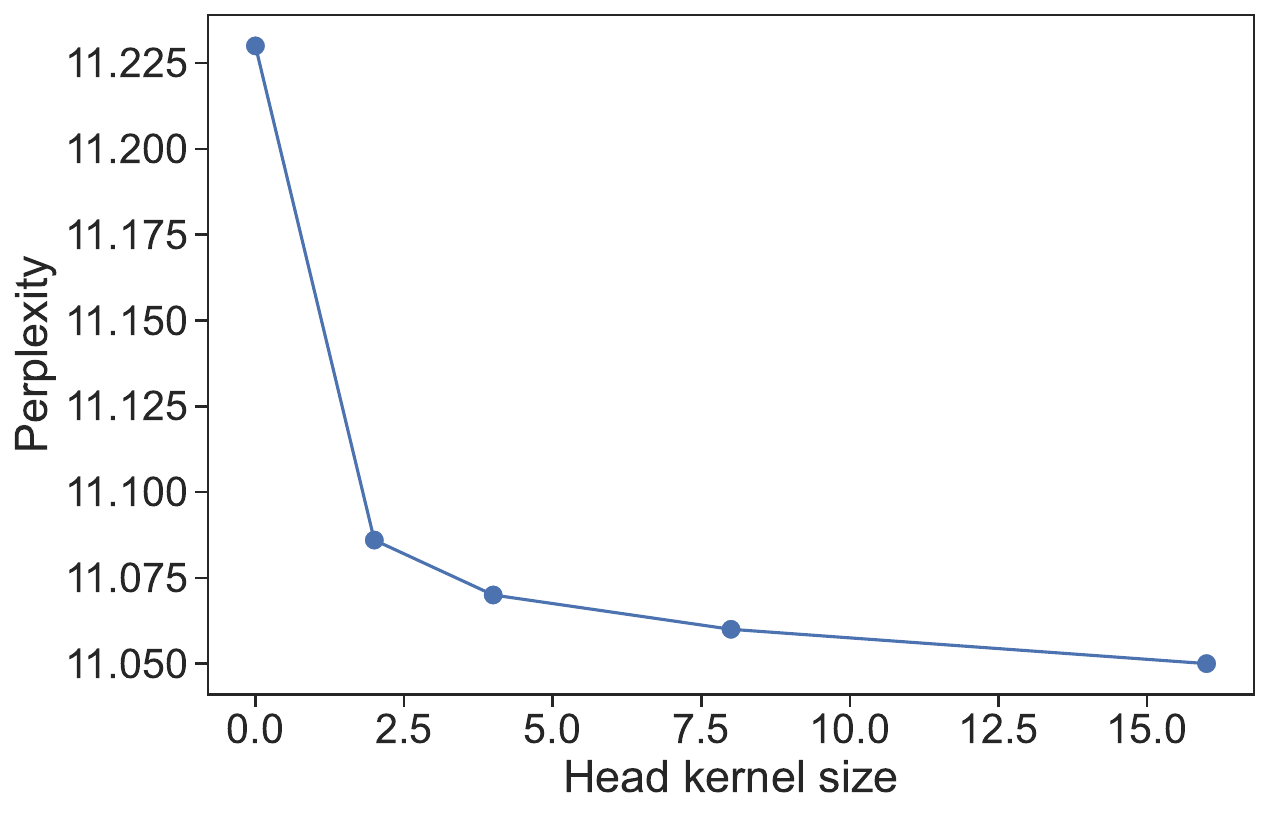}
    \hfill
    \includegraphics[height=5.3cm]{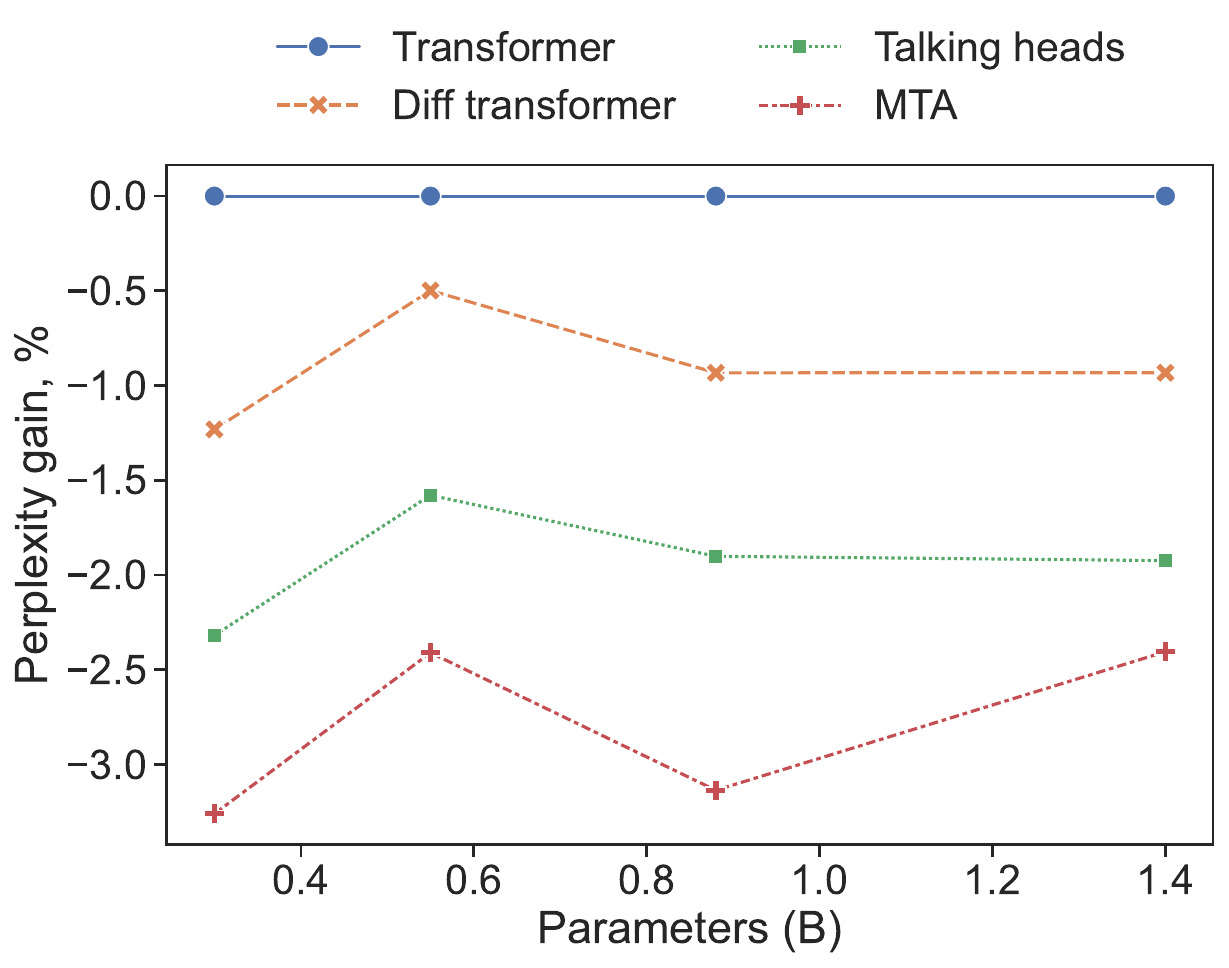}
    \hfill
    \caption{ Ablation on the head kernel size in \ours{} convolution \textbf{(left)}, and scaling laws \textbf{(right)}. We report average validation perplexity on SlimPajama.}
    \label{fig:head_abl}
\end{figure}

\paragraph{Head convolution} An ablation study on the head kernel size shows that increasing kernel size improves average validation perplexity, as shown in \autoref{fig:head_abl} (left), allowing for better communication across heads.

\paragraph{Kernel initialization} Kernel initialization can affect the convergence rate and stability of training. We evaluated \ours{} models initialized with zero, constant ($0.3$), and identity kernels. The latter corresponds to initialization with a regular Transformer without convolution in the attention. We found that identity initialization leads to better convergence and final performance, while models initialized with zero and $0.3$ values reduce average validation perplexity by $0.02$ and $0.08$ correspondingly. 

\paragraph{Ablation on \ours{} components}
We further experiment with \ours{} kernels of different sizes, changing the order of convolution operations with respect to softmax, and different normalizations, such as layer-norm scaling \citep{sun2025curse}. Results are summarized in \autoref{tab:abl-valid-ppl}. We found that different kernel sizes display similar kernel patterns, while resulting in slightly different evaluation results (middle rows). Group normalization and exponential depth scaling are both important factors each bringing improvement over vanilla \ours{} (top rows), while square root layer-norm scaling surprisingly underperforms. Finally, changing the order of convolutions wrt softmax operations increases perplexity only by 0.01-0.04 points (bottom rows).

\begin{table*}[t!]
\footnotesize
  \centering
\begin{tabular}{l|l|l|l|l|c}
\toprule
\multicolumn{2}{c|}{\bf Key-query conv} & \multicolumn{2}{c|}{\bf Head conv} & \\
\bf Pre-sm & \bf Post-sm  & \bf Pre-sm & \bf Post-sm  & \bf Group norm & \bf PPL $\downarrow$ \\
\midrule
$\checkmark$ & $\checkmark$ & $\checkmark$ & $\checkmark$ & scalar gating & \bf 10.90 \\
$\checkmark$ & $\checkmark$ & $\checkmark$ & $\checkmark$ & depth scaling & 10.92 \\
\midrule
$\checkmark$ & $\times$ & $\checkmark$ & $\checkmark$ & scalar gating & 10.95 \\
$\checkmark$ & $\times$ & $\checkmark$ & $\checkmark$ & $\times$ &  10.99 \\
$\checkmark$ & $\times$ & $\checkmark$ & $\checkmark$ & depth scaling & 10.99 \\
$\checkmark$ & $\times$ & $\checkmark$ & $\checkmark$ & no scaling & 11.03 \\
\midrule
$\checkmark$ & $\times$ & $\times$ & $\checkmark$ & depth scaling &11.09 \\
$\checkmark$ & $\times$ & $\times$ & $\checkmark$ & $\times$ & 11.16 \\
$\checkmark$ & $\times$ & $\times$ & $\checkmark$ & no scaling & 11.13 \\
$\checkmark$ & $\times$ & $\times$ & $\checkmark$ & layer-norm scaling & 11.41 \\
\midrule
$c_q=4$, $c_k=9$ & $\times$ & $\times$ & $\times$ & depth scaling &  11.23 \\
$c_q=6$, $c_k=11$ & $\times$ & $\times$ & $\times$ & depth scaling & 11.23 \\
$c_q=8$, $c_k=13$ & $\times$ & $\times$ & $\times$ & depth scaling & 11.31 \\
\midrule
 $\times$ & $\checkmark$ & $\times$ & $\checkmark$ & depth scaling & 11.11 \\
 $\times$ & $\checkmark$ & $\times$ & $\checkmark$ & $\times$ & 11.19 \\
$\checkmark$ & $\times$ & $\checkmark$ & $\times$ & depth scaling & 11.10 \\
$\checkmark$ & $\times$ & $\checkmark$ & $\times$ & $\times$ & 11.20 \\
\bottomrule
\end{tabular}

\captionof{table}{Ablation on \ours{} components: validation perplexity over SlimPajama dataset.}
\label{tab:abl-valid-ppl}
\end{table*}

\paragraph{Scaling laws} We experiment with models of various sizes: 300M, 550M, and 1B, with hyperparameters listed in \autoref{tab:hyperparam}. As shown in \autoref{fig:head_abl} (right), we observe a consistent pattern across models of different sizes, where MTA provides superior performance. 

\section{Related work}
\paragraph{Focusing attention} There has been a number of attempts to focus the soft attention mechanism on important context. 
For example, modifications have been proposed to make softmax sharper. \cite{martins2016softmax} propose to replace softmax with sparsemax for sparse activations. \cite{velivckovic2024softmax,nakanishi2025scalable} propose Adaptive temperature and Scalable-Softmax that adjust exponential base in softmax attention.
Irrelevant tokens can be altogether removed from memory \citep{Sukhbaatar2021NotAM} to make attention easier.
In \citet{golovneva2024contextual,desrochers2024reducing} the authors incorporate contextual information into position encodings to allow positions to be amplified by context. 
More recently, several noise canceling mechanisms  have been proposed for attention. Talking heads attention \citep{shazeer2020talking} adds linear projections across the attention-heads dimension before and after the softmax operation.  \textsc{Diff} transformer \citep{ye2024differential} uses differential amplifiers to focus attention to the relevant context, which is related to our head mixing step. 
\citet{cang2025dint} further extends it by introducing normalization steps.
OpAmp adaptation \citep{wu2025efficient} propose  adapters, which are fine-tuned with noisy context to enhance an LLMs’ denoising capability. 
\citet{Xu2024KVSA} modified attention so that keys and values can shift by one time step. This can be viewed as a special case of \ours{} where the convolution performs shifting in the key dimension.

\paragraph{Convolution in Attention}
Convolution layers have been used with attention \citep{gehring2017convolutional,wu2019pay}, especially in vision Transformers, to  decompose each image into a sequence of tokens with fixed length, and then
apply multiple standard Transformer layers \citep{xiao2021early, wu2021cvt}. 
There have been some attempts to include convolution in language modeling as well. For example, \cite{liu2018generating} use convolution to \textit{compress} the keys and values in the multi-headed
attention by a factor of 3 thus allowing to process sequences 3x in length. \cite{liu2019hierarchical} later augment this architecture with the ability to encode multiple documents in a hierarchical manner, and further improve on long-context summarization tasks. Later \citet{subramanian2020multi} propose three hierarchical models to learn different levels of abstraction in language by interchanging casual transformer layers with convolutions or pooling, or compressing representations for tokens further away in the past. 
\cite{gulati2020conformer} proposed a Conformer architecture for speech recognition, where a convolution module is applied to multi-head self-attention output. 
This architecture was later used to encode speech in the Llama-3 herd of models \citep{grattafiori2024llama}. \citep{zheng2024dape} apply a convolution on the attention weights in the key dimension that enhances Transformer’s extrapolation capabilities to long context.
However, none of these methods apply convolutions across multiple queries, keys and heads to the attention weights. 

\section{Conclusion}
In this paper, we focused on a limitation of the standard soft attention mechanism that stems from conditioning on the similarity of a single vector pairs.
This makes it challenging for Transformers to precisely locate relevant information based on richer distinguishing information.
As a remedy, we proposed a novel \oursfull{} mechanism that combines attention weights from multiple queries, keys and heads, making it possible to attend using more fine-grained information.
From a simple motivating toy task to large-scale LLM experiments on a variety of popular benchmarks we demonstrate that models equipped with \ours{} achieve enhanced performance, especially when tested on tasks that require the precise location of relevant information within a long context.




\bibliography{colm2025_conference}
\bibliographystyle{colm2025_conference}

\appendix

\section{Limitations}
To the best of our knowledge, the \oursfull{} method is not currently compatible with the popular optimized attention kernels. Optimizing the runtime performance was not the goal of this work, thus we leave further optimization of the MTA implementation for future research. 

\section{\ours{} Architecture}
\paragraph{Head convolution vs normal attention with higher rank}
Let us consider mixing of two heads after softmax.
If $W^1_o$ and $W^2_o$ are the output projections corresponding to those heads,
their output can written as 
\begin{align*}
O &= (A^1_\text{new} V^1) W^1_o + (A^2_\text{new} V^2) W^2_o\\
 &= (w_{11} A^1 V^1 + w_{12} A^2 V^1) W^1_o + ( w_{21} A^1 V^2 + w_{22} A^2 V^2) W^2_o \\
 &=  A^1 (w_{11} V^1 W^1_o + w_{21} V^2 W^2_o) + A^2 (w_{12} V^1 W^1_o + w_{22} V^2 W^2_o)  \\
 &=  A^1 (w_{11} H W^1_v W^1_o + w_{21} H W^2_v W^2_o) + A^2 (w_{12} H W^1_v W^1_o + w_{22} H W^2_v W^2_o)  \\
 &=  A^1 H (w_{11} W^1_v W^1_o + w_{21} W^2_v W^2_o) + A^2 H (w_{12} W^1_v W^1_o + w_{22} W^2_v W^2_o) 
\end{align*}
If we compare the first term to a normal attention output  $A^1 H W^1_v W^1_o $, we note that the only difference is that now we have two rank-$d$ matrix additions instead of one. Actually we can rewrite
\[
w_{11} W^1_v W^1_o + w_{21} W^2_v W^2_o = \hat{W}_1^v \hat{W}_1^o
\]
where $\hat{W}^1_v \md{D}{2d}$ and $\hat{W}^1_o \md{2d}{D}$. Thus, post-softmax head convolution can be replicated by a normal attention head with twice the rank. 
This may help in understanding why head mixing is useful, as it may increase the expressive power using a higher rank.
However, it is not truly identical to $2\times$ rank because the parameters of the two heads are not independent.

Let us see if the same is true for the pre-softmax version. The attention logits after mixing is
\begin{align*}
 \hat{A}^1 &= w_{11} Q^1 K^{1\top} + w_{12} Q^2 K^{2\top} \\ &=  w_{11} H W^1_q W_k^{1 \top} H^\top + w_{12} H W^2_q W_k^{2 \top} H^\top \\
 &= H (w_{11} W^1_q W_k^{1 \top} + w_{12} W^2_q W_k^{2 \top}) H^\top
\end{align*}
Again, this can be rewritten like a normal attention logits
\[ \hat{A}^1 = H \hat{W}^1_q \hat{W}_k^{1 \top} H^\top\]
where $\hat{W}^1_q, \hat{W}^1_k \md{D}{2d}$.
As before, it can be replicated by a normal attention with twice the rank in the key and query projections.

\section{Toy task details}
\label{sec:toy_detail}
The blocks are built by randomly choosing $N \in \{5, 8\}$ lowercase alphabets and concatenating them. We join up to 50 blocks into a single sequence separated by ``.'', followed by ``\#'' and $L=2$ question letters. Here is an example sequence: 

\begin{center}
\texttt{hjnvt.qfjgt.whftb.bjtpq.} \emph{...(many blocks)...} \texttt{.pxjvf.ulhik.qoiax\#pb}    
\end{center}
Here the 4th block ``\texttt{bjtpq}'' is the target because it contains all query letters ``\texttt{pb}''.
During data generation, we make sure there is only one target block in each sequence. We use 1M such sequences for training and test on held-out 1K sequences.

We train a small Transformer model with 4 layers, 2 heads and 256 hidden dimensions.
The batch size is 64 and the training continues for a total of 100K steps.
We run each experiment three times with different random seeds and report the average performance.

\begin{table*}[ht]
  \centering
  \small
\begin{tabular}{l|l|l|l|l}
\toprule
Task &  Name  & Input & Target & Question\\
\midrule
QA1 & single supporting fact & \makecell[l]{John travelled to \\the hallway. \\Mary journeyed \\to the bathroom. \\Daniel went back \\to the bathroom. \\John moved to \\the bedroom.} & bathroom & Where is Mary?\\
\midrule
QA2 & two supporting fact & \makecell[l]{Mary journeyed to \\the bathroom. ... \\Daniel grabbed the \\football there. \\Sandra grabbed the \\milk there. \\Daniel went to \\the kitchen.} & kitchen & Where is the football?\\
\midrule
QA3 & three supporting fact & \makecell[l]{Daniel moved to \\the bathroom. ...\\ Mary got the football. \\Mary went back to\\ the kitchen. \\Mary journeyed to\\ the garden.} & kitchen & \makecell[l]{Where was the football\\ before the garden?}\\
\midrule
QA4 & two arg relations & \makecell[l]{The hallway is east \\ of the bathroom. \\The bedroom is \\west of the bathroom.} & bedroom & \makecell[l]{What is the \\bathroom east of?}\\
\midrule
QA5 & three arg relations & \makecell[l]{Fred picked up the \\football there. \\Fred gave the \\football to Jeff. ... \\Jeff gave the \\football to Fred. \\Fred gave the \\football to Jeff.} & Jeff & \makecell[l]{Who did Fred give \\the football to?}\\
\bottomrule
\end{tabular}
\caption{Examples of QA1-5 tasks in BabiLong benchmark when there is no distraction context. }
\label{tab:babilong_examples}
\end{table*}

\begin{figure}

    \centering
    \includegraphics[width=0.32\linewidth]{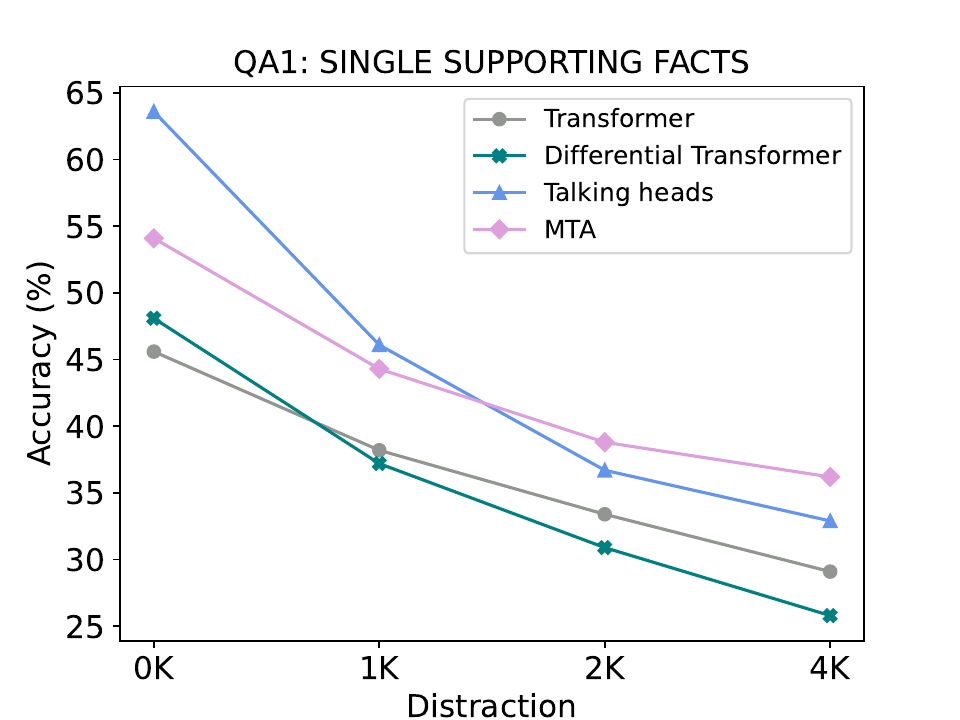}
    \includegraphics[width=0.32\linewidth]{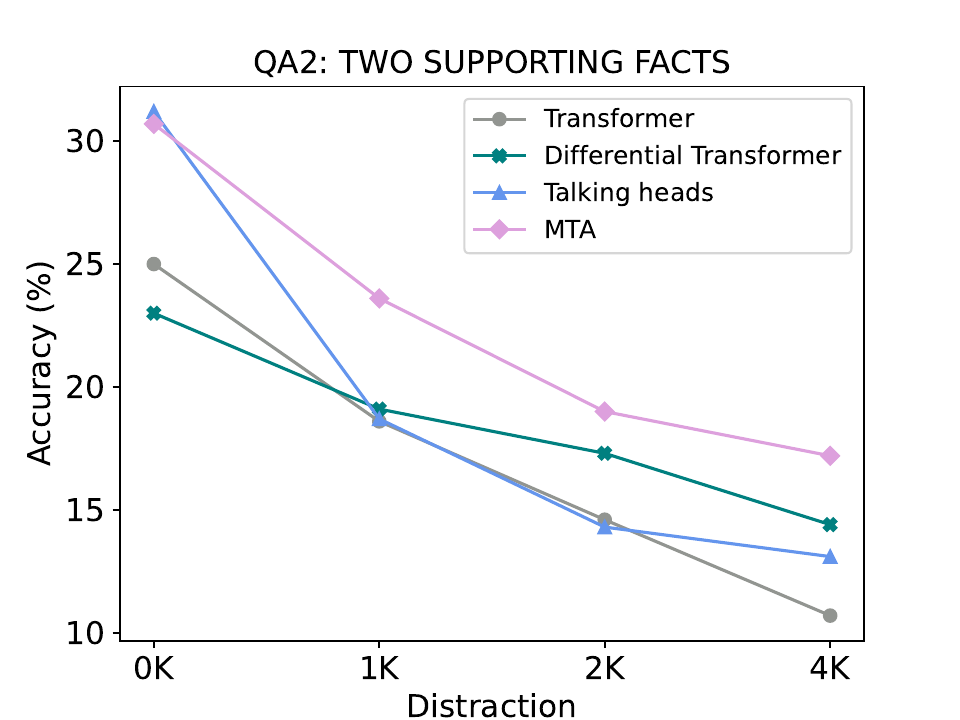}
    \includegraphics[width=0.32\linewidth]{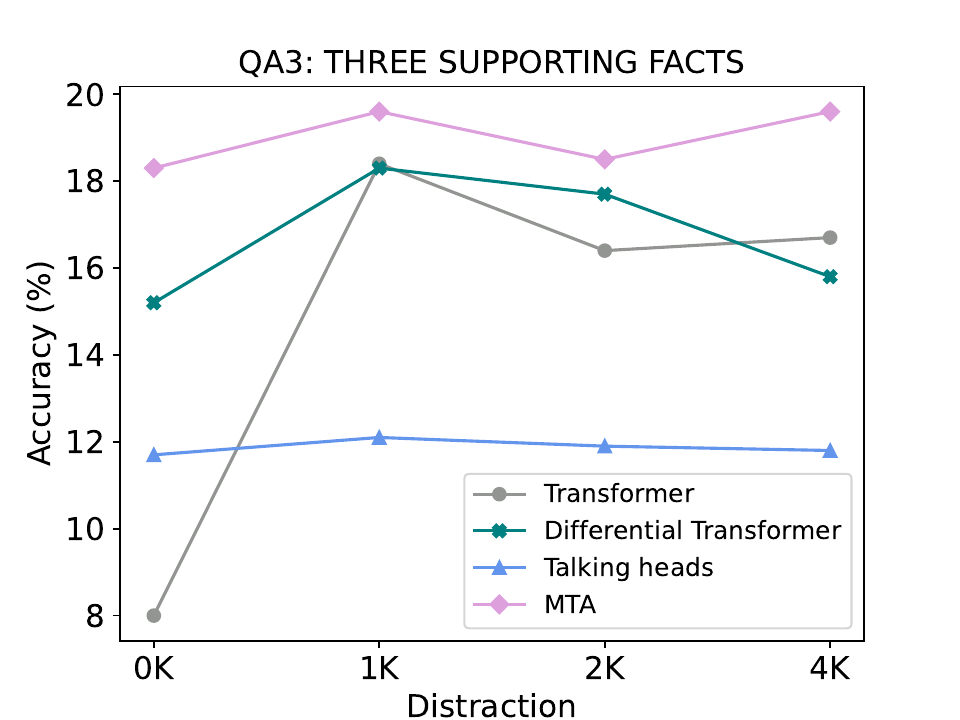}
    \includegraphics[width=0.32\linewidth]{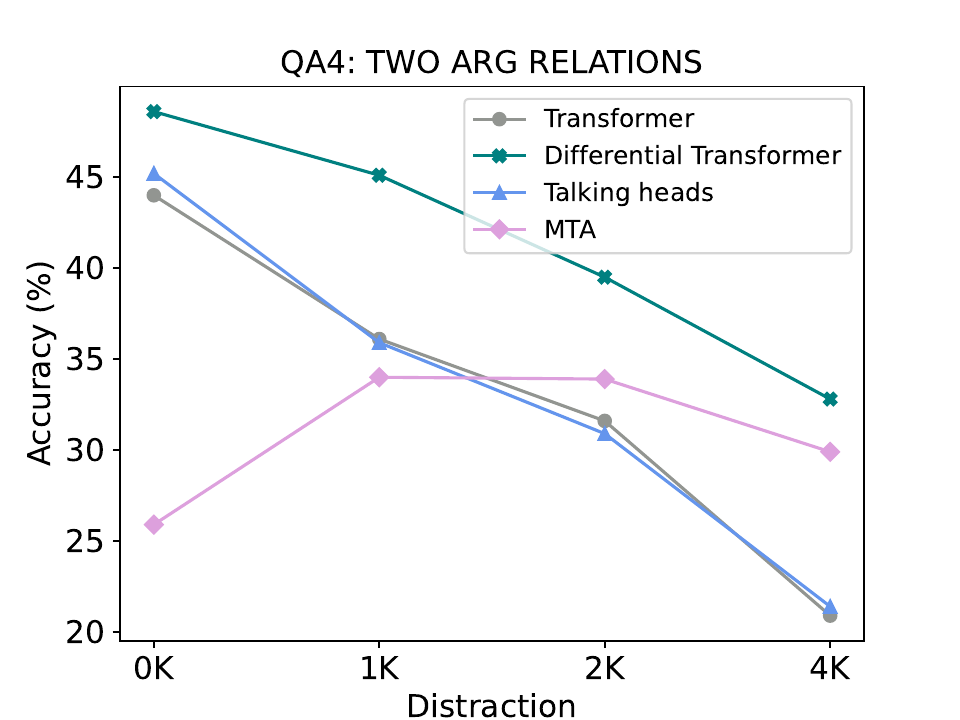}
    \includegraphics[width=0.32\linewidth]{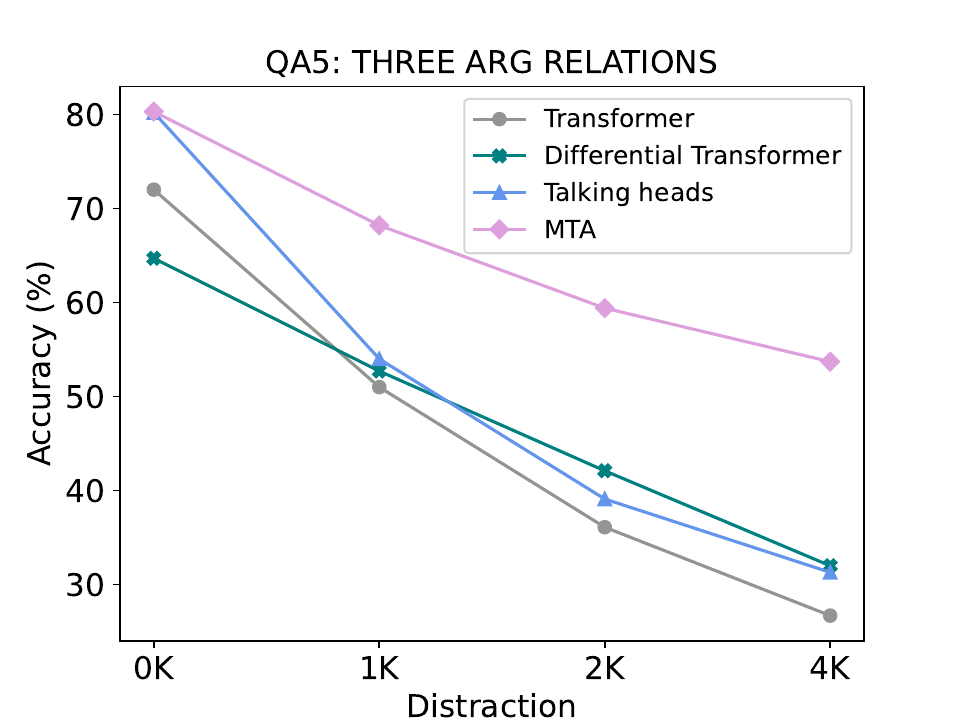}
    \caption{Accuracy (\%) on QA1-5 tasks in BabiLong benchmark. Note the random performance on QA3 is 16.67\%, thus all models perform poorly on QA3. On other tasks, MTA demonstrates strong performance compared to baselines especially when there is lengthy distraction text (4K).}
    \label{fig:qa1_3}
\end{figure}

\section{Language model training details}
\label{sec:trainig_details}
Training setup follows LLaMa architecture \citep{touvron2023llama}, and includes RMSNorm pre-normalization \citep{zhang2019root}, SwiGLU activation function \citep{shazeer2020glu}, and Rotary Embeddings \citep{su2024roformer}.
\autoref{tab:hyperparam} shows a detailed hyperparameter breakdown for Transformer model pretraining. All models were trained in the same setup, except Diff Transformer, which had twice less heads to account for doubled dimension.

We evaluate on the following tasks: BoolQ~\citep{clark2019boolq}, PIQA~\citep{bisk2020piqa}, SIQA~\citep{sap2019socialiqa}, HellaSwag~\citep{zellers2019hellaswag}, WinoGrande~\citep{sakaguchi2021winogrande}, ARC easy and challenge~\citep{clark2018think}, OpenBookQA~\citep{mihaylov2018can}. We also report 5-shot performance on the aggregated MMLU benchmark~\citep{hendrycks2020measuring}.

\begin{table*}[t!]
  \centering
\begin{tabular}{l|cccc}
\toprule
\bf Parameter & \bf 300M model & \bf 550M model & \bf 880M model & \bf 1B model \\
\midrule
Dimension, $D$ & $1024$ & $1280$ & $1536$ & $2048$ \\
Layers & $20$ & $24$ & $24$ & $24$ \\
Heads, $M$ & $16$ & $10$ & $16$ & $16$ \\
RoPE theta & $100,000$ & $100,000$ & $100,000$ & $100,000$ \\
Batch size (tokens) & $262,144$ & $262,144$ & $262,144$ & $524,288$ \\
Context length & $2048$ & $2048$ & $2048$ & $2048$ \\
Learning rate & $1.5e-4$& $1.5e-4$& $1.5e-4$& $1.5e-4$ \\
Weight decay & $0.05$ & $0.05$ & $0.05$ & $0.05$ \\
Warm up steps & $375$ & $375$ & $375$ & $187$ \\
\bottomrule
\end{tabular}
\caption{Hyperparameters for Transformer training.}
\label{tab:hyperparam}
\end{table*}

\section{BabiLong details}
We include examples of input and output from BabiLong benchmark in \autoref{tab:babilong_examples}. The details of per-task evaluation results on QA1-5 are shown in Figure~\ref{fig:qa1_3}.

\section{Complexity evaluation}
\label{sec:compute}

\autoref{tab:extra-param} shows estimated additional number of parameters and their actual counts. We note that \textsc{Diff} transformer \citep{ye2024differential} introduces four learnable vectors per layer ($\lambda_{q1},\lambda_{k1},\lambda_{q2},\lambda_{k2}$), plus group normalization weights. In total, that results in more parameters than \oursfull{} with key-query convolution added to the quarter of layers, even though all kernel weights in \ours{} models are different between all heads and layers.

\autoref{tab:mem-and-runtime} shows actual runtime and memory consumption recorded during 880M model training.
Note that the baseline and \textsc{Diff} transformer utilize \textit{scaled dot product attention}\footnote{\hyperlink{1}{https://pytorch.org/docs/stable/generated/torch.nn.functional.scaled\_dot\_product\_attention.html}} function implemented in Torch, that calls optimized CUDA kernels for improved performance.
In contrast, our \ours{} implementation does not take advantage of such efficient kernels, which is the major reason behind its lower FLOPS.

\begin{table*}[t!]
 \small
  \centering
\begin{tabular}{lcc}
\toprule
Model & Additional parameters, estimates
& Actual number of parameters \\
\midrule
Transformer &  & 876,553,728 \\
\textsc{Diff} transformer & $4 * L*d + 2 * L * d$ & 876,567,552 \\
Talking heads & $2 * L * M * M$ & 876,566,016 \\
MTA & $2 * L* M(c_q*c_k/4 + c_h) + L*(2*d+1)$ & 876,583,320 \\
\bottomrule
\end{tabular}
\caption{Estimation of the parameter counts for each model architecture.}
\label{tab:extra-param}
\end{table*}

\begin{table*}[t!]
 \small
  \centering
\begin{tabular}{lccc}
\toprule
Model & Memory, GB $\downarrow$
& FLOPS, $10^{13}\uparrow$  & TPS, $10^{3}\uparrow$ \\
\midrule
Transformer & 17.5 & 25.0 & 54.3 \\
\textsc{Diff} transformer & 21.6 & 8.4 & 17.6 \\
Talking heads & 63.0 & 6.9 & 14.6 \\
MTA & 73.8 & 2.6 & 5.7 \\
\bottomrule
\end{tabular}
\caption{Average memory consumption and training speed for models trained on 32 NVIDIA H200 GPUs with our unoptimized code.}
\label{tab:mem-and-runtime}
\end{table*}

\section{Multi-needle evaluation results}
\label{app:needle}

\begin{figure}
    \centering
    \includegraphics[width=0.48\linewidth]{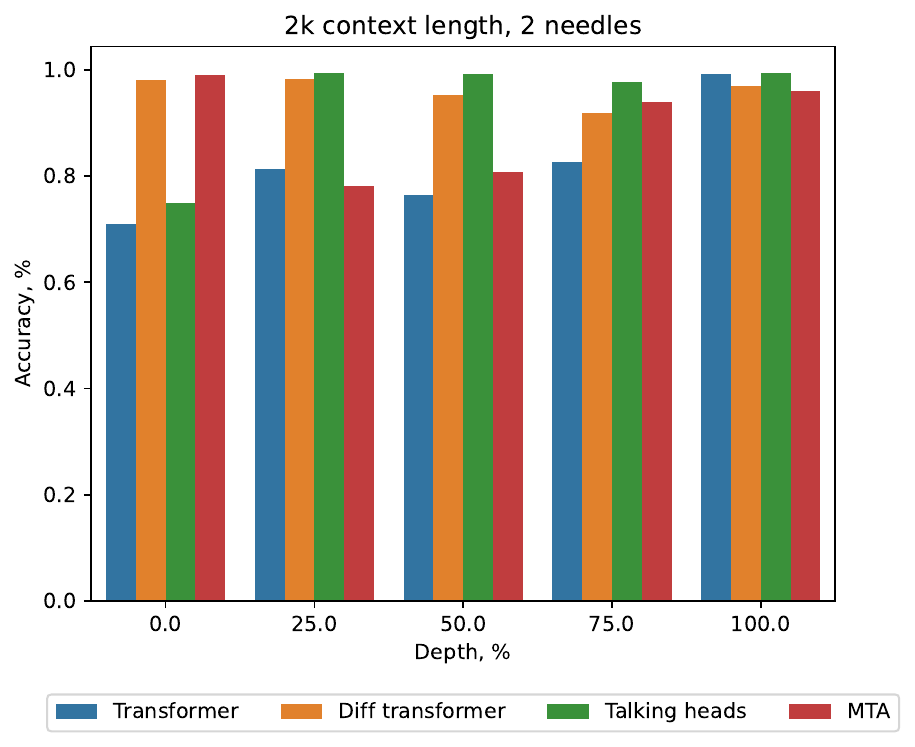}\includegraphics[width=0.48\linewidth]{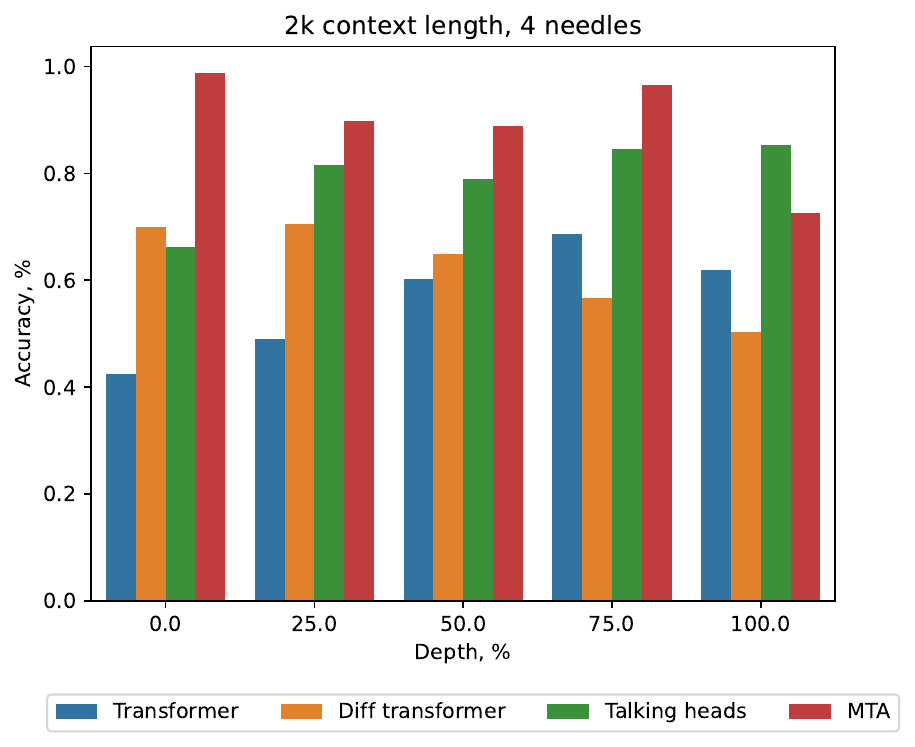}
    \includegraphics[width=0.48\linewidth]{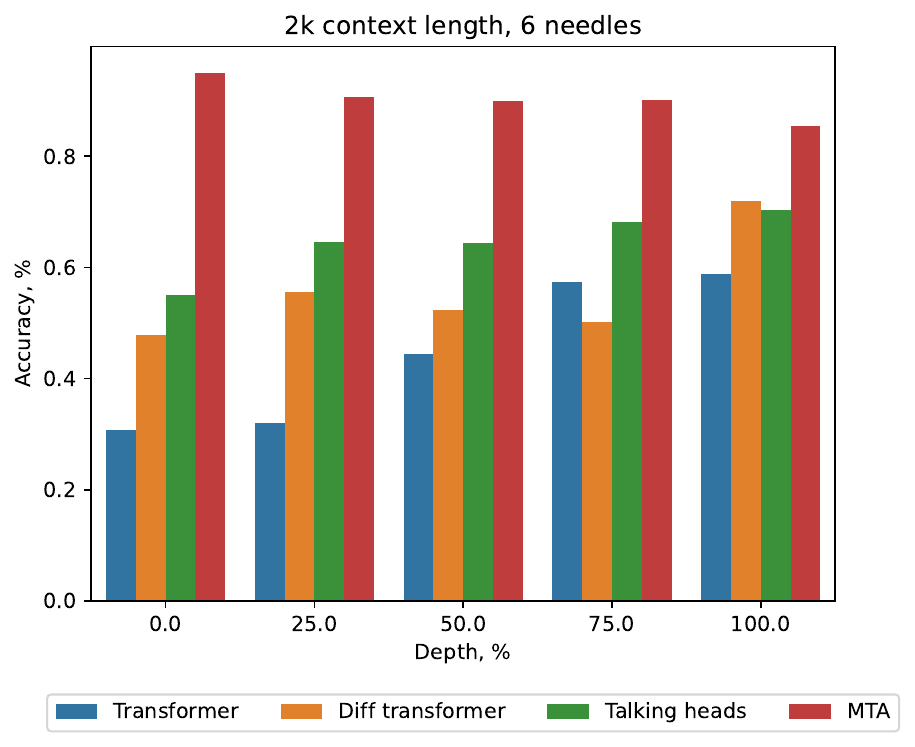}\includegraphics[width=0.48\linewidth]{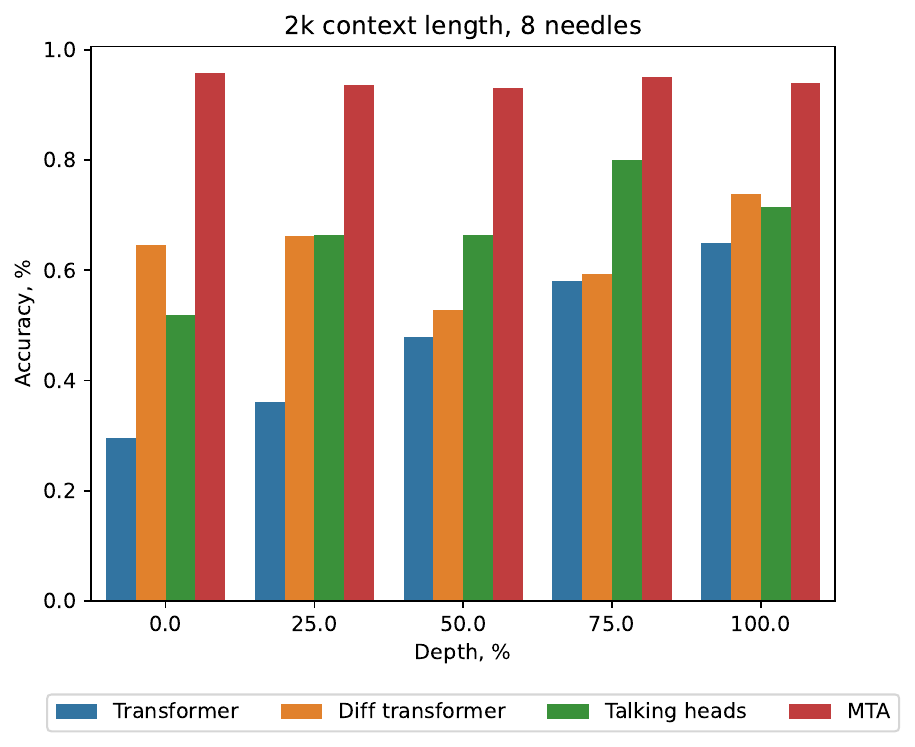}
    \caption{Needle-in-a-Haystack accuracy across different models. Models trained with \ours{} are better at predicting needles hidden deep in the context.}
    \label{fig:needle-2k}
\end{figure}

We report on the detailed evaluation results in \autoref{fig:needle-2k}. We observe that models trained with \ours{} are better at finding needles hidden deeper in the context. 

\section{\ours{} kernel patterns.}
\label{app:kernel_patterns}

\begin{figure}
    \centering
    \includegraphics[width=1\linewidth]{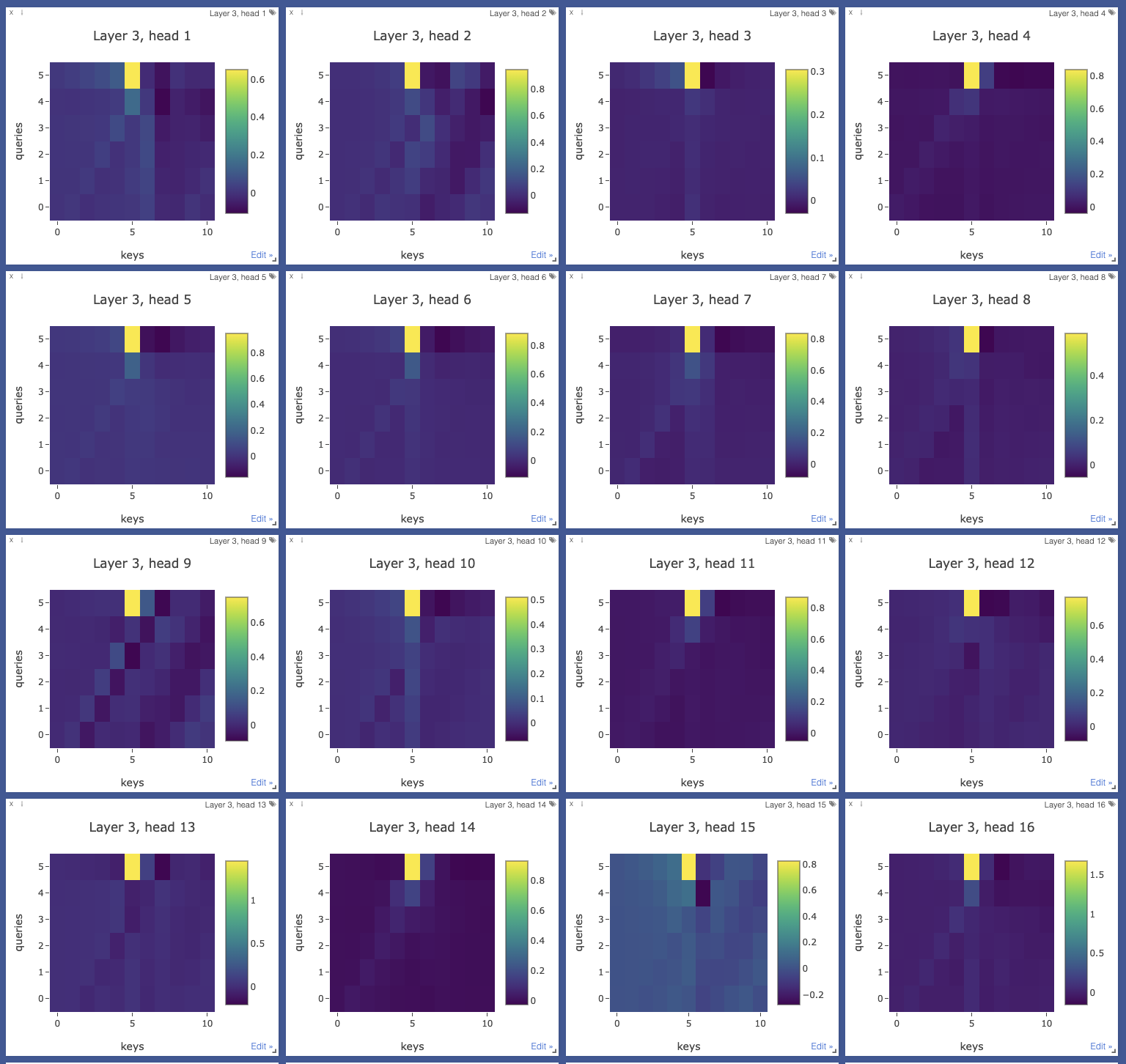}
    \caption{Key-query kernel patterns across heads at 3rd layer of the Transformer model with \ours{}.}
    \label{fig:kern_l3}
\end{figure}

\begin{figure}
    \centering
    \includegraphics[width=1\linewidth]{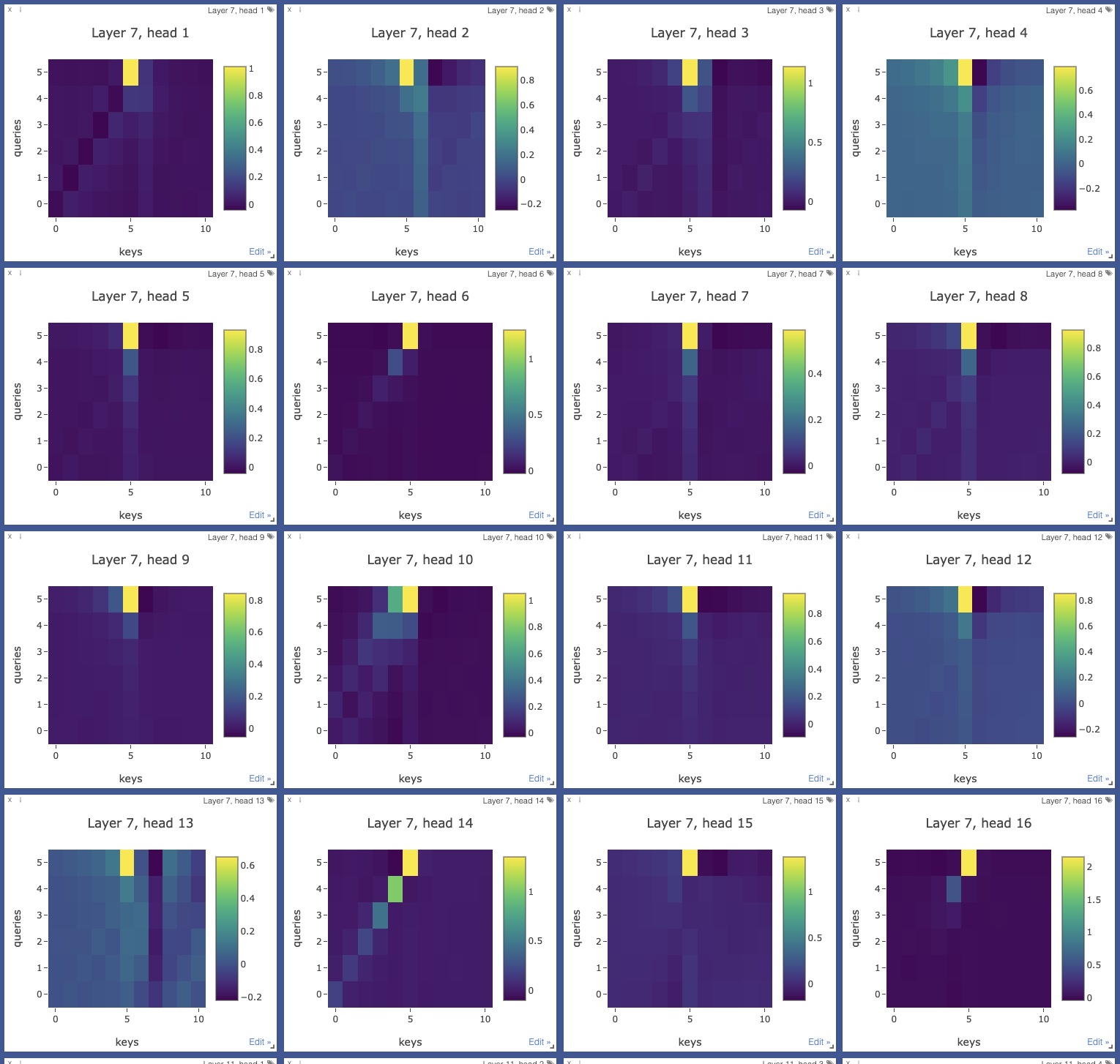}
    \caption{Key-query kernel patterns across heads at 7th layer of the Transformer model with \ours{}.}
    \label{fig:kern_l7}
\end{figure}

\begin{figure}
    \centering
    \includegraphics[width=1\linewidth]{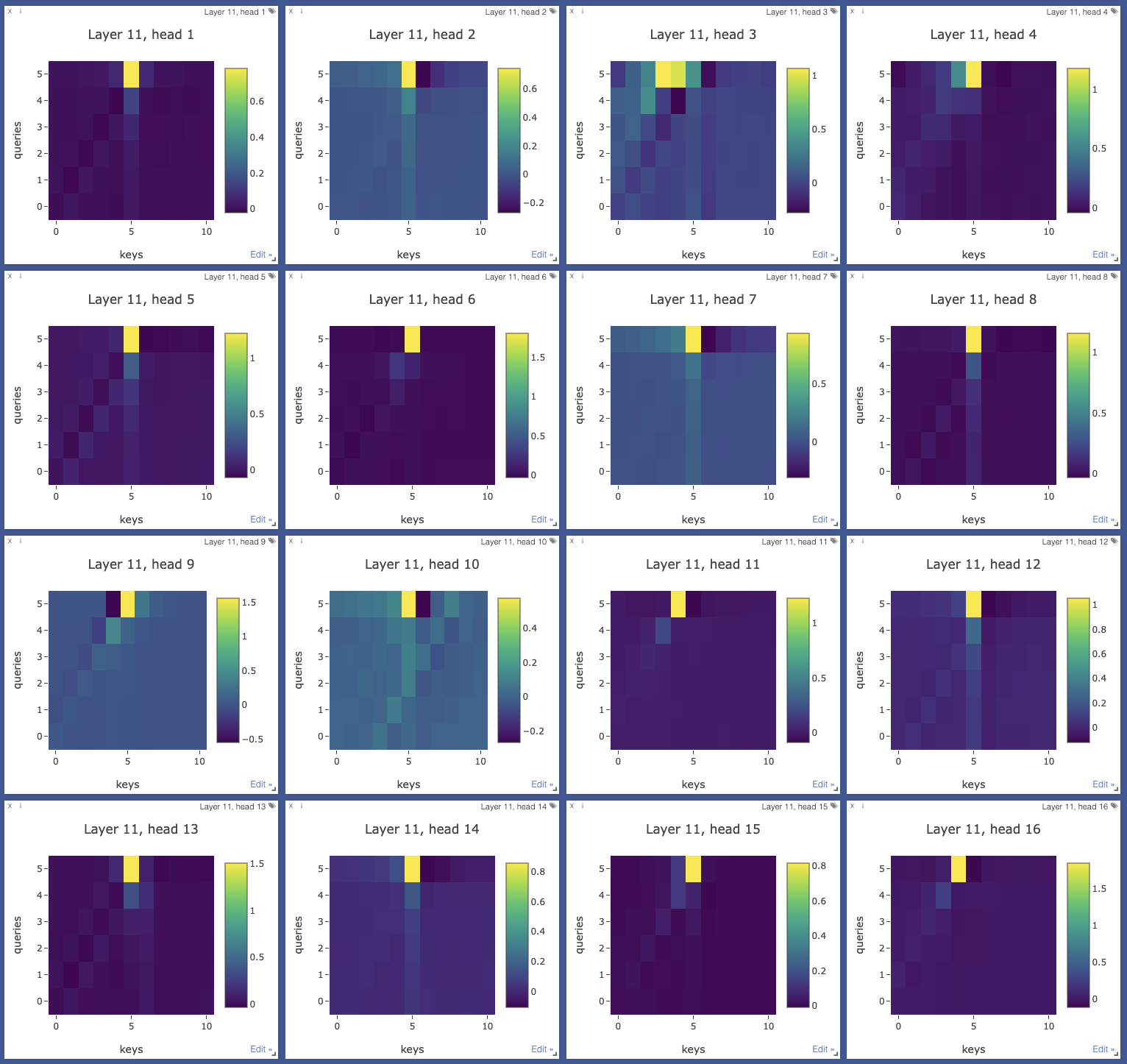}
    \caption{Key-query kernel patterns across heads at 11th layer of the Transformer model with \ours{}.}
    \label{fig:kern_l11}
\end{figure}

\begin{figure}
    \centering
    \includegraphics[width=1\linewidth]{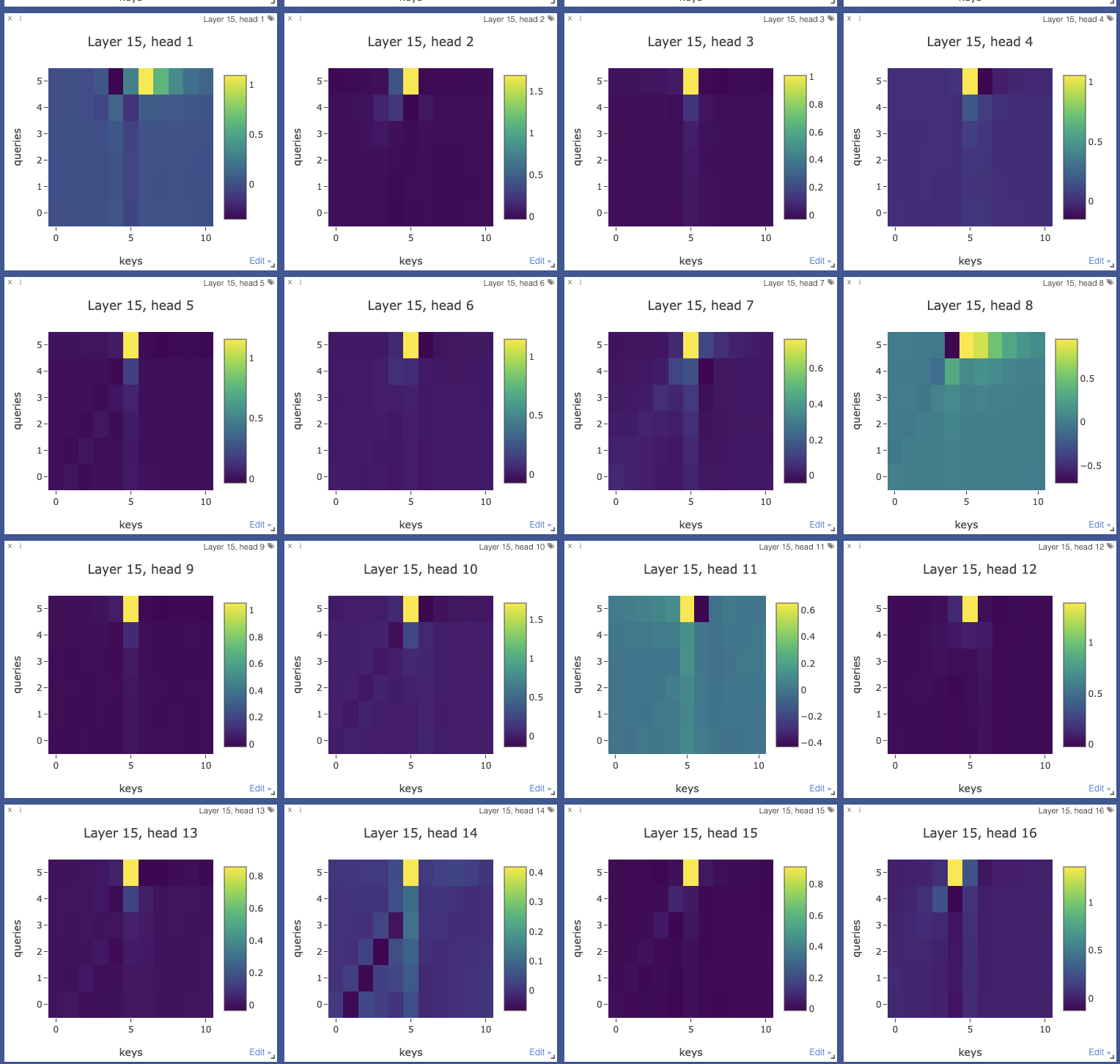}
    \caption{Key-query kernel patterns across heads at 15th layer of the Transformer model with \ours{}.}
    \label{fig:kern_l15}
\end{figure}

\begin{figure}
    \centering
    \includegraphics[width=1\linewidth]{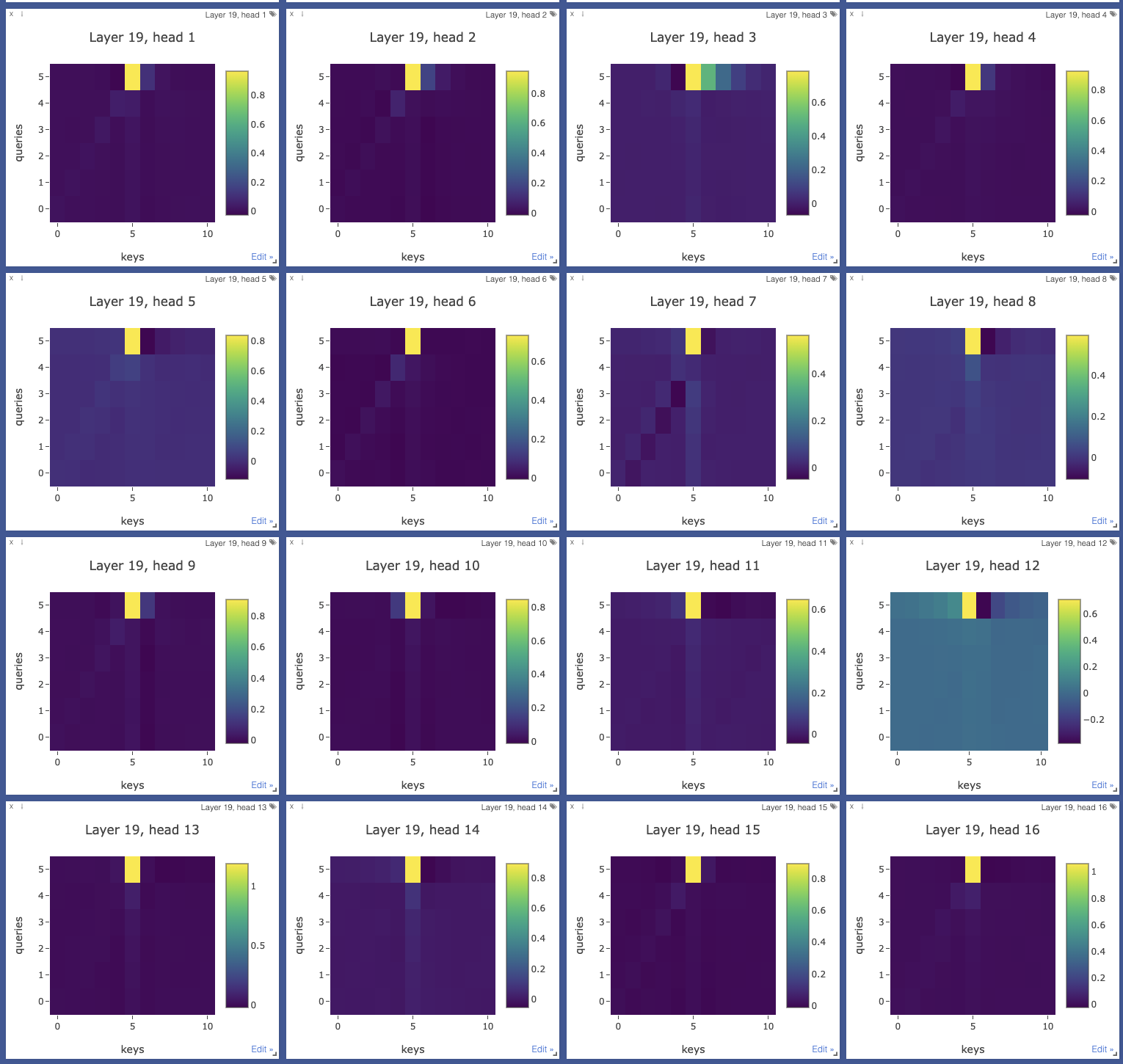}
    \caption{Key-query kernel patterns across heads at 19th layer of the Transformer model with \ours{}.}
    \label{fig:kern_l19}
\end{figure}

\begin{figure}
    \centering
    \includegraphics[width=1\linewidth]{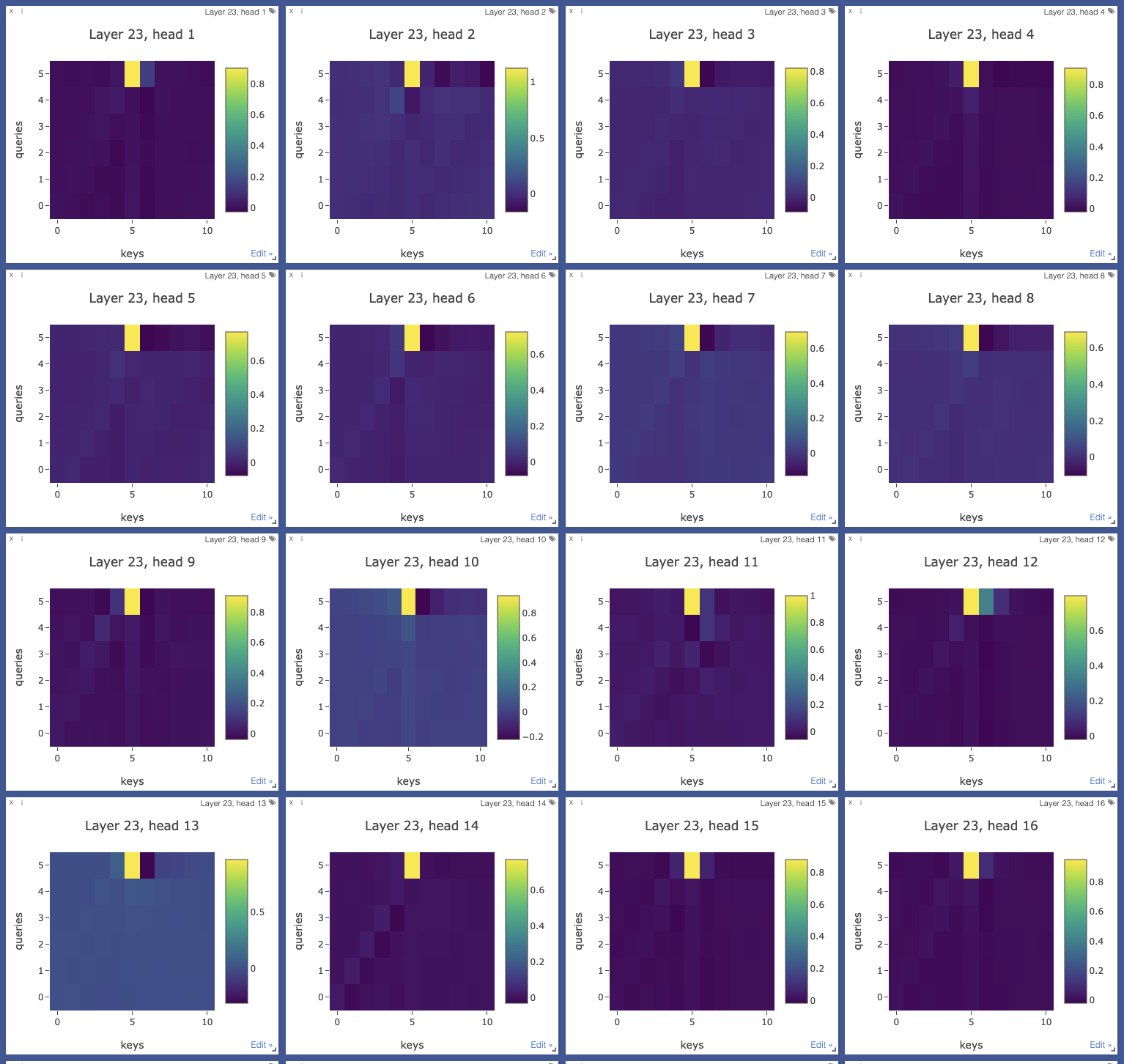}
    \caption{Key-query kernel patterns across heads at 23rd layer of the Transformer model with \ours{}.}
    \label{fig:kern_l23}
\end{figure}

Key-query convolution kernels across different layers and heads are reported in \autoref{fig:kern_l3}-\autoref{fig:kern_l23}. 

\begin{figure}
    \centering
    \includegraphics[width=1\linewidth]{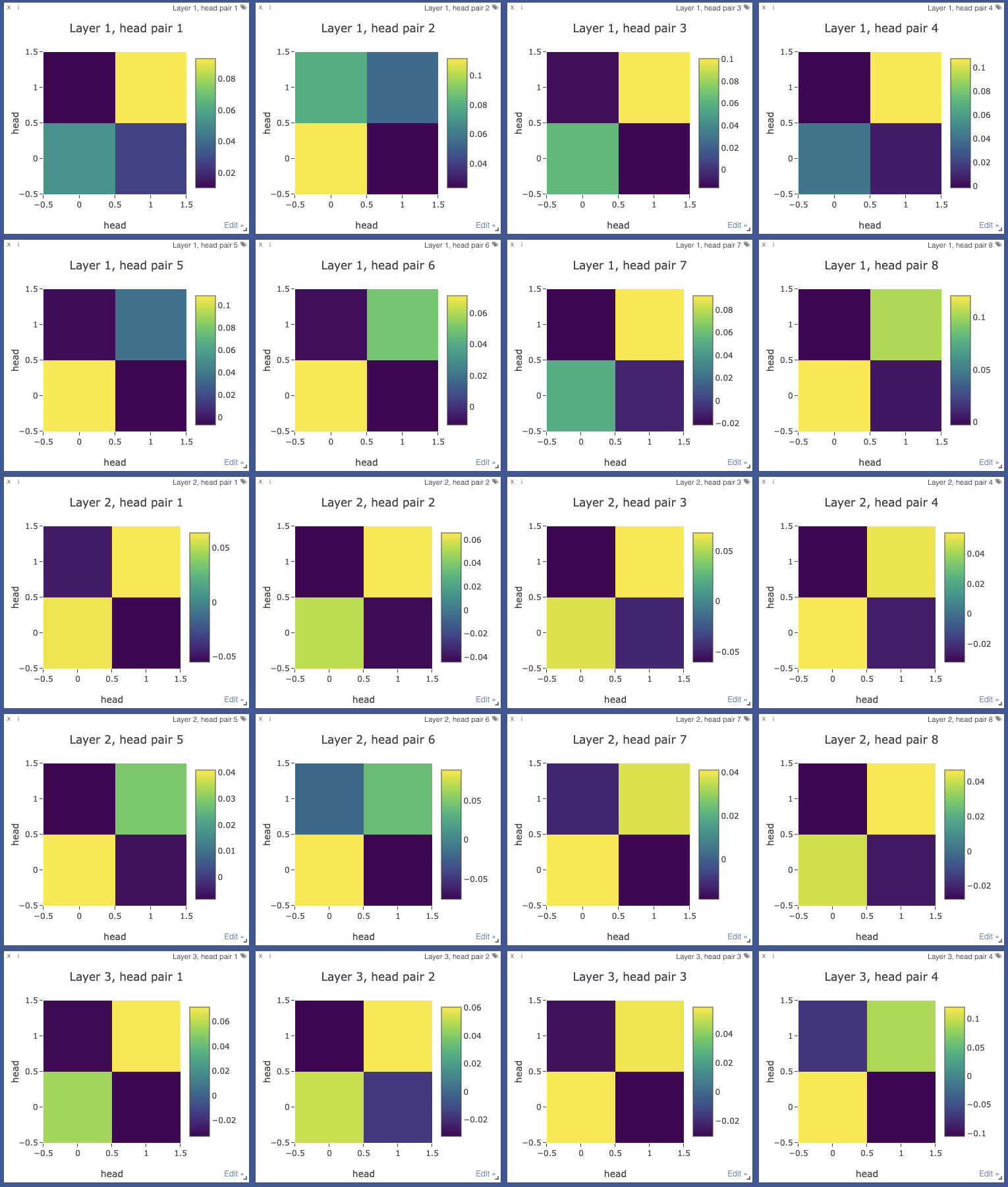}
    \caption{Head kernel patterns across first three layers of the Transformer model with \ours{}.}
    \label{fig:head_kern}
\end{figure}

\begin{figure}
    \centering
    \includegraphics[width=0.45\linewidth]{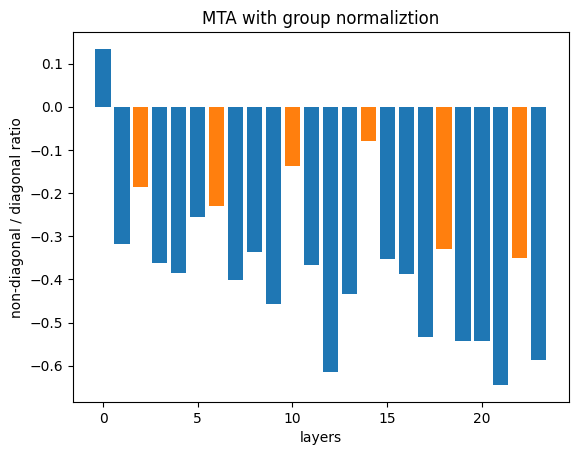}
    \includegraphics[width=0.45\linewidth]{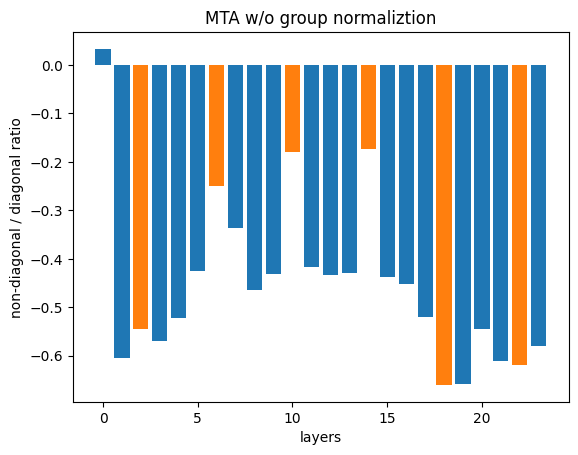}
    \caption{Average of the ratio between non-diagonal and diagonal kernel values across head pairs for \oursfull{} models with (left) and without (right) group normalization. Values for layers with key-query convolution are colored in orange.}
    \label{fig:head_ratio}
\end{figure}

Head kernel patterns for the model with $c_h$=2 are shown in \autoref{fig:head_kern}. They are simpler due to their small size. Besides an identity with scaling, a common pattern is contrasting:
subtracting one attention weight from another. We also observe that the kernel scales
increase with layers when there is no group normalization (see \autoref{fig:head_ratio}). This is
probably to compete with the residual stream, which gets larger with the model’s depth.
However, this pattern is not present with group normalization because it will undo the
effect of such scaling.

\begin{figure}
    \centering
    \includegraphics[width=0.45\linewidth]{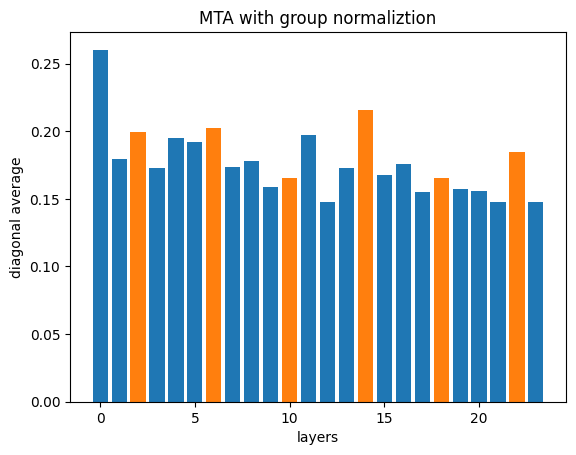}
    \includegraphics[width=0.45\linewidth]{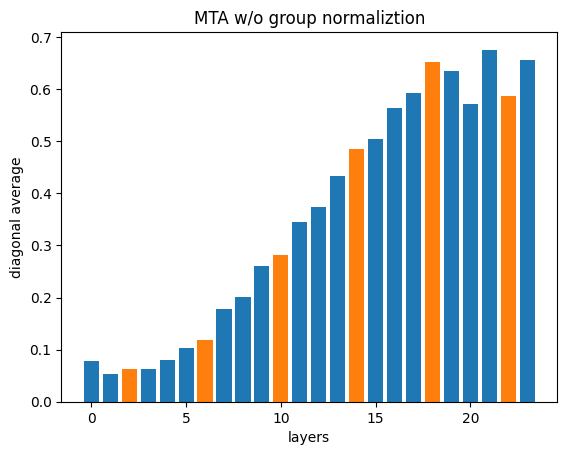}
    \caption{Average of the diagonal kernel values across head pairs for \oursfull{} models with (left) and without (right) group normalization. Values for layers with key-query convolution are colored in orange.}
    \label{fig:head_diag}
\end{figure}

\section{Finetuning with \ours{}}

One natural question readers might ask is if \ours{} could be integrated into models that were already trained with standard attention? Would this require complete retraining, or could it be added through some form of adaptation?
Architectually, \ours{} can be added as additional layers to already trained models, and weights can be updated with continual training. Since the main component of \ours{} is a convolution, we can initialize it to identity and insert it in an existing a Transformer layer. Such a modification will not change the output of the layer, so when we start training the model, it should maintain its performance. However, as the added convolution starts deviating from the identity state, it will allow the transformer to condition its attention on multiple keys and queries. 

To understand how well pre-trained models can learn we perform preliminary experiments by finetuning our 1.4B model, as well as opensourced Llama models \citep{grattafiori2024llama}. In all these experiments we used the same finetuning setup similar to \autoref{sec:long-cont}, but kept the context length at 2048 tokens. 

Validation perplexity results are reported in \autoref{tab:ft-valid-ppl}. We observe that all models finetuned with \ours{} were not only able to incorporate new kernels, but outperform baselines in terms of perplexity. 

\begin{table*}[t]
  \centering
  \small
\begin{tabular}{ll|cccccccc}
\toprule
\bf Pretraining & \bf Cont training & \bf arxiv & \bf book & \bf c4 & \bf cc & \bf github & \bf se & \bf wiki & \bf Avg PPL $\downarrow$ \\
\midrule
\hspace{-2mm}\textit{1.4B models} \\
Transformer & Transformer & 4.54 & 12.87 & 19.21 & 13.77 & 3.99 & 9.49 & 10.96 & 10.69 \\
Transformer & \ours{} & 4.51 & 12.78 & 19.09 & 13.67 & 3.95 & 9.41 & 10.87 & 10.61 \\
\ours{} & \ours{} & \bf 4.47 & \bf 12.57 & \bf 18.82 & \bf 13.47 & \bf 3.89 & \bf 9.25 & \bf 10.65 & \bf 10.44 \\
\midrule
\hspace{-2mm}\textit{Llama 3 herd} \\
Llama 3.2 1B & Llama 3.2 1B & 4.54 & 14.60 & 18.14 & 13.40 & 3.92 & 8.84 & 10.25 & 10.53\\
Llama 3.2 1B & \ours{} & 4.52 & 14.49 & 18.04 & 13.32 & 3.89 & 8.79 & 10.19 & 10.46 \\
Llama 3.2 3B & Llama 3.2 3B & 4.12 & 12.08 & 15.58 & 11.50 & 3.37 & 7.52 & 8.35 & 8.93 \\
Llama 3.2 3B & \ours{} & 4.11 & 12.03 & 15.51 & 11.46 & 3.35 & 7.48 & 8.31 & 8.89 \\
Llama 3.1 8B & Llama 3.1 B & 4.00 & 11.04 & 15.04 & 10.99 & 3.29 & 7.33 & 8.00 & 8.53 \\
Llama 3.1 8B & \ours{} & \bf 3.98 & \bf 10.97 & \bf 14.97 & \bf 10.94 & \bf 3.27 & \bf 7.29 & \bf 7.97 & \bf 8.48 \\
\bottomrule
\end{tabular}
\caption{Validation perplexity on SlimPajama dataset after continuous training for 10.5B tokens on our 1.4B models, and continuous training for 5.3B tokens for Llama 3 models.}
\label{tab:ft-valid-ppl}
\end{table*}

\end{document}